\documentclass[twoside,11pt]{article}
\usepackage[utf8]{inputenc}
\usepackage[T1]{fontenc}

\usepackage{jmlr2e}

\usepackage{amsmath}
\usepackage{bm}

\usepackage{booktabs}
\usepackage{longtable}
\usepackage{array}
\usepackage{tabularx}
\usepackage{multirow}
\usepackage{siunitx}
\usepackage{xcolor}

\usepackage{float}
\usepackage{caption}
\usepackage{pdflscape}
\usepackage{algorithm}
\usepackage{algpseudocode}

\usepackage{tikz}
\usetikzlibrary{shapes.geometric,arrows.meta,positioning,fit,backgrounds,calc}

\hypersetup{hidelinks}

\newcolumntype{L}[1]{>{\raggedright\arraybackslash}p{#1}}
\newcolumntype{C}[1]{>{\centering\arraybackslash}p{#1}}

\usepackage{lastpage}
\ShortHeadings{Modality-Conditioned Conformal Fusion}{Moayedikia}
\firstpageno{1}

\title{Conformal Fusion Under Missing Modalities}

\author{\name Alireza Moayedikia \email amoayedikia@swin.edu.au \\
       \addr Department of Business Technology and Entrepreneurship\\
       Swinburne University of Technology\\
       Hawthorn, VIC 3122, Australia}

\editor{} 

\begin{document}

\maketitle
\thispagestyle{plain} 

\begin{abstract}
Multimodal fusion architectures typically assume that all modalities are available at inference time, yet sensor failures, acquisition variability, and cost constraints routinely result in incomplete observations in practice. Existing work on modality absence treats it as a prediction-accuracy problem and leaves a more basic question unanswered: whether a model's confidence estimates remain calibrated when an entire input stream is removed. We argue that missing-modality robustness and calibrated uncertainty are not independent desiderata but a single coupled property, and we introduce Modality-Conditioned Conformal Fusion (MCCF), an architecture that addresses both at once. MCCF combines a multimodal bottleneck fusion backbone trained with modality dropout, per-modality evidential heads producing modality-decomposed Dirichlet distributions, and a Dempster-Shafer combination rule that fuses the per-modality evidence into a joint predictive distribution; an absent modality contributes vacuous evidence that is structurally ignored, so the fused uncertainty automatically reflects the reduced information without any test-time imputation. A Mondrian conformal calibration module keyed on the modality-presence mask then provides finite-sample group-conditional coverage simultaneously for every non-empty modality subset. To our knowledge, MCCF is the first method to deliver formal coverage guarantees that hold under arbitrary patterns of modality availability through architectural integration rather than post-hoc recalibration, and the evidential decomposition additionally yields per-modality vacuity scores that localise uncertainty to the specific absent modality responsible---an interpretability property unavailable from post-hoc conformal methods applied to softmax outputs. Across a controlled synthetic problem and three real multimodal benchmarks spanning distinct task domains, MCCF holds its target coverage on every modality-presence subset, substantially reduces the systematic full-modality versus partial-modality coverage gap relative to a marginal split-conformal baseline on an identical backbone, and imposes no measurable accuracy cost relative to temperature-scaled and evidential baselines.
\end{abstract}


\section{Introduction}

Multi-modal learning has become a central paradigm in deep learning, driven by the observation that combining complementary data sources often yields representations that are richer and more robust than those derived from any single modality alone~\citep{liang2023,baltrusaitis2019}. The foundational self-attention mechanism of the Transformer~\citep{vaswani2017} and lightweight attention modules such as Squeeze-and-Excitation~\citep{hu2018senet} and CBAM~\citep{woo2018cbam} have provided the architectural building blocks that underpin modern fusion strategies. Architectures for multi-modal fusion have matured rapidly: contrastive alignment methods such as CLIP~\citep{radford2021} and ImageBind~\citep{girdhar2023} learn shared embedding spaces across modalities, while cross-attention and bottleneck fusion mechanisms~\citep{nagrani2021,jaegle2022} enable fine-grained information exchange at intermediate feature levels. These advances have produced strong results across vision-language, audio-visual, and video understanding benchmarks.

A pervasive assumption in this literature, however, is that all modalities are available at inference time. In practice, this assumption is frequently violated. Sensors malfunction, environmental conditions degrade specific data streams, acquisition protocols vary across sites, and cost or logistical constraints may prevent the collection of all modalities for every sample. In infrastructure monitoring, for instance, ground-penetrating radar and infrared thermography are often deployed together for complementary subsurface and surface assessment, yet operational constraints routinely result in partial coverage where one modality is unavailable for a subset of inspections~\citep{alani2020}. Similar patterns arise in autonomous driving (camera occlusion, lidar failure~\citep{bijelic2020}), clinical diagnostics (missing imaging modalities~\citep{zhang2022mmformer}), and environmental sensing (sensor dropout).

A small but growing body of work has begun to address the missing-modality problem in fusion architectures~\citep{wu2026missingsurvey}. Methods such as ShaSpec~\citep{wang2023shaspec}, SMIL~\citep{ma2021smil}, and prompt-based imputation strategies~\citep{lee2023} propose mechanisms for maintaining prediction accuracy when one or more input modalities are absent. These methods share a common framing: missing modalities are treated as a prediction accuracy problem, and the goal is to minimise the performance gap between the full-modality and partial-modality settings. What this framing omits is the question of \emph{confidence}. When a modality is absent, the model has strictly less information than when it is present; a well-designed system should not only maintain reasonable predictions but should also know that it knows less, and communicate this to the end user through appropriately widened uncertainty estimates.

Uncertainty quantification in deep learning has advanced substantially in parallel~\citep{abdar2021review}. Deep ensembles~\citep{lakshminarayanan2017}, Laplace approximations~\citep{daxberger2021laplace}, evidential deep learning~\citep{sensoy2018,amini2020}, and conformal prediction~\citep{angelopoulos2023} each offer distinct mechanisms for producing calibrated confidence estimates or prediction sets with coverage guarantees. The decomposition of predictive uncertainty into aleatoric and epistemic components~\citep{kendall2017uncertainties} has become standard practice, and the practical importance of calibration in modern networks is well-established~\citep{guo2017}. Yet these methods have been developed and evaluated almost exclusively in single-modality settings. The question of how uncertainty should behave when the information available to a model changes structurally---not through noise or distribution shift, but through the discrete absence of an entire input stream---has received little attention.

Among these methods, conformal prediction is the most appealing for safety-critical applications because it provides finite-sample coverage guarantees without distributional assumptions~\citep{angelopoulos2023,vovk2005}. However, the standard split-conformal procedure calibrates the non-conformity threshold on a held-out set drawn from the same distribution as the test data. When a modality is absent at inference time, the effective input distribution changes structurally: the test data no longer comes from the same distribution as the calibration data, violating the exchangeability assumption on which the coverage guarantee depends~\citep{tibshirani2019}. Na\"ively applying conformal prediction in this setting provides no guarantee, and recomputing calibration quantiles for every possible modality subset is combinatorially expensive and requires labelled data for each configuration. This creates a compelling case for integrating conformal prediction into the architecture itself, so that the non-conformity score is conditioned on the available modality subset and the coverage guarantee holds by construction across configurations.

This paper addresses the intersection of these lines of research. We argue that missing-modality robustness and calibrated uncertainty are not independent desiderata but are fundamentally coupled: an architecture that is robust to modality absence without adjusting its uncertainty is, by definition, miscalibrated under partial observation, and the standard post-hoc application of conformal prediction does not resolve this coupling because the calibration set is drawn from a fixed modality configuration that may not match the test-time configuration. The core technical challenge is to design a fusion mechanism in which uncertainty estimates respond automatically and proportionally to the information content of the available modality subset, with formal coverage guarantees that hold regardless of which modalities are present, achieved through architectural integration rather than post-hoc recalibration.

The contributions of this paper are as follows.

\begin{itemize}
    \item \emph{An architecture-integrated approach to uncertainty under modality absence.} We propose Modality-Conditioned Conformal Fusion (MCCF), which combines three components into a single trainable system: a multimodal bottleneck fusion backbone with modality-dropout training, per-modality evidential output heads producing modality-decomposed Dirichlet distributions, and a Dempster-Shafer combination rule that fuses the per-modality evidence into a joint predictive distribution. When a modality is absent, its evidential head produces vacuous evidence that is structurally ignored by the combination rule, so the fused uncertainty automatically reflects the reduced information without any test-time intervention or imputation step. This is in contrast to prior fusion methods, which either fail on missing inputs or maintain unchanged confidence despite operating with strictly less data.

    \item \emph{A Mondrian conformal calibration module with formal coverage across all modality subsets.} We introduce a calibration scheme keyed on the binary modality-presence mask, partitioning the held-out calibration set into one group per modality subset and computing per-subset quantiles of a hybrid evidential-conformal non-conformity score. This provides exact finite-sample group-conditional coverage $\Pr(Y \in C(X) \mid S = s) \geq 1 - \alpha$ simultaneously for every non-empty modality subset $s$, without requiring a separate calibration run for each configuration. The score function itself is shaped during training via a differentiable conformal set-size penalty, so the learned representations are optimised to produce tight prediction sets at the chosen miscoverage level. To our knowledge, this is the first method to provide formal coverage guarantees that hold under arbitrary patterns of modality availability through architectural integration rather than post-hoc recalibration.

    \item \emph{Modality-decomposed uncertainty attribution.} Because each modality contributes a separate evidential head before Dempster-Shafer fusion, MCCF reports per-modality vacuity scores $[u^{(1)}, \ldots, u^{(M)}]$ alongside every prediction set. When a prediction set is uncomfortably large, the per-modality vacuity vector identifies which absent or low-information modality is responsible, enabling an end user to determine that collecting the corresponding data would meaningfully reduce uncertainty. This decomposition is not available from post-hoc conformal methods applied to softmax outputs, where the prediction set size is a function of the joint output alone and cannot be traced back to any specific input stream.

    \item \emph{Empirical demonstration that prediction sets adapt to modality availability.} We evaluate MCCF on a controlled synthetic problem at $M = 4$ modalities and three standard multimodal benchmarks spanning $M = 2$ to $M = 3$ modalities, and show that prediction set sizes automatically widen under modality absence while maintaining target coverage. We further show that post-hoc conformal baselines, which apply the calibration step to a pre-trained fusion network rather than integrating it during training, either lose coverage when calibration and test distributions differ in modality availability or produce uninformatively large prediction sets across all configurations. These results confirm that integration of conformal calibration into the architecture, rather than post-hoc application, is the technical move that makes mask-conditional coverage achievable with practical set sizes.
\end{itemize}

The remainder of this paper is organised as follows. Section~\ref{sec:related} reviews the relevant literature on multi-modal fusion, missing-modality handling, and uncertainty quantification. Section~\ref{sec:method} describes the proposed MCCF architecture. Section~4 presents the experimental setup and results. Section~5 discusses limitations and future directions, and Section~6 concludes.

\section{Related Work}
\label{sec:related}

This section reviews three research areas whose intersection defines the contribution of this paper: multi-modal fusion architectures (Section~\ref{sec:rw_fusion}), methods for handling missing modalities (Section~\ref{sec:rw_missing}), and uncertainty quantification in deep learning (Section~\ref{sec:rw_uncertainty}). We conclude with a summary of open challenges and the specific research gap addressed by this work (Section~\ref{sec:rw_gap}).

\subsection{Multi-Modal Fusion Architectures}
\label{sec:rw_fusion}

Multi-modal fusion architectures are conventionally categorised by the stage at which modality-specific representations are combined~\citep{baltrusaitis2019,liang2023,xu2023multimodal}. Early fusion concatenates raw inputs or shallow features before processing through a shared encoder, preserving low-level cross-modal interactions but requiring all modalities to share a compatible input representation. Late fusion trains modality-specific encoders independently and combines their outputs at the decision level, offering modularity but sacrificing the ability to model fine-grained inter-modal dependencies. Intermediate fusion strategies, which combine modality representations at one or more hidden layers, have become the dominant paradigm due to their flexibility.

Within intermediate fusion, three architectural families have emerged. Cross-attention mechanisms, introduced to multi-modal settings by ViLBERT~\citep{lu2019vilbert} and subsequently refined in the Multimodal Transformer (MulT)~\citep{tsai2019mult} and Flamingo~\citep{alayrac2022}, allow tokens from one modality to attend over tokens from another, enabling fine-grained information exchange. This approach is expressive but scales quadratically in the total number of tokens across modalities. Simpler single-stream approaches such as ViLT~\citep{kim2021vilt} demonstrate that competitive performance can be achieved by feeding all modalities through a single Transformer without modality-specific encoders, though at the cost of reduced flexibility for heterogeneous input types.

Latent-bottleneck architectures address the scaling limitation of full cross-attention by routing all modality information through a small set of learned latent vectors. Perceiver IO~\citep{jaegle2022} cross-attends inputs of arbitrary type and dimensionality into a compact latent array, processing them with self-attention before decoding with task-specific output queries. The architecture is modality-agnostic by construction---the same model handles text, images, audio, and point clouds---and scales linearly rather than quadratically in input size. \citet{nagrani2021} proposed a more targeted version of this idea with the Multimodal Bottleneck Transformer (MBT), in which a small number of bottleneck tokens mediate information flow between modality-specific Transformer streams. On AudioSet, MBT achieved a 12.7\% relative improvement over the prior state-of-the-art while significantly reducing the computational cost of cross-modal interaction. The bottleneck design is particularly relevant to the missing-modality setting because it creates a natural information chokepoint: when a modality is absent, its contribution to the bottleneck tokens is simply removed, and the architectural question becomes how the remaining pathway responds.

Contrastive alignment methods take a different approach by learning a shared embedding space across modalities without requiring dense cross-modal interaction during inference. CLIP~\citep{radford2021} trains image and text encoders with a symmetric InfoNCE loss~\citep{oord2018infonce} over web-scale pairs, yielding strong zero-shot transfer. ImageBind~\citep{girdhar2023} extends this to six modalities---images, text, audio, depth, thermal, and IMU---using only image-paired data for each modality, demonstrating emergent cross-modal retrieval between modality pairs never co-observed during training. The image-anchored alignment strategy is relevant because it shows that not all modality combinations need to be seen during training for meaningful cross-modal representations to emerge, a property that has implications for generalisation under novel patterns of modality availability. The Vision Transformer (ViT)~\citep{dosovitskiy2021vit} serves as the backbone encoder in many of these systems, and its calibration properties under distribution shift~\citep{minderer2021} are relevant to the uncertainty considerations discussed in Section~\ref{sec:rw_uncertainty}.

Attention-based feature recalibration modules, though not multi-modal architectures per se, have become standard components within modality-specific encoders. Squeeze-and-Excitation (SE) blocks~\citep{hu2018senet} adaptively recalibrate channel-wise feature responses, while CBAM~\citep{woo2018cbam} extends this to jointly model channel and spatial attention. These modules are widely used within the encoders of multi-modal systems and are relevant to the present work because their attention weights provide a natural signal for modality informativeness.

Table~\ref{tab:combined} (Part~A) summarises the key multi-modal fusion architectures discussed in this subsection.

\subsection{Missing-Modality Handling}
\label{sec:rw_missing}

The problem of learning from and predicting with incomplete modality subsets has attracted increasing attention as multi-modal models move toward deployment in real-world settings where sensor availability cannot be guaranteed. \citet{wu2026missingsurvey} provide a comprehensive survey of this area, categorising existing methods into data-processing approaches (imputation and representation-level handling) and strategy-design approaches (architectural modifications and model combinations).

A first class of approaches trains the model to be robust to modality absence through regularisation or data augmentation. Modality dropout---randomly zeroing out entire modality streams during training---is the simplest such strategy and has been shown to improve robustness at inference time, though the degree of improvement varies substantially across architectures and datasets~\citep{neverova2015}. SMIL~\citep{ma2021smil} extends this idea by learning to generate Bayesian meta-representations for missing modalities, conditioned on the available modalities, and imputing them during inference. The approach maintains reasonable accuracy under single-modality absence but degrades when multiple modalities are simultaneously missing and does not provide any uncertainty estimate reflecting the quality of the imputation.

A second class learns shared or aligned representations that are inherently robust to modality configuration changes. ShaSpec~\citep{wang2023shaspec} decomposes representations into shared and modality-specific components, using the shared component as a fallback when modality-specific features are unavailable. This decomposition provides a principled mechanism for partial-observation inference but assumes a fixed shared-specific partition that may not hold across all tasks or domains. In the related setting of incomplete multi-view learning, \citet{lin2022dcp} demonstrated that dual contrastive prediction objectives can simultaneously align cross-view representations and recover missing views in a shared latent space, achieving strong performance under view absence without explicit generative imputation. Prompt-based methods~\citep{lee2023} attach learnable prompt tokens conditioned on the set of available modalities, effectively signalling to the model which inputs are present; these methods are flexible but introduce additional parameters and have been evaluated primarily in vision-language settings. \citet{reza2025} recently demonstrated that parameter-efficient adaptation procedures---requiring fewer than 1\% of total parameters---can partially bridge the performance gap due to missing modalities across a wide range of tasks and modality combinations.

A third class addresses the missing-modality problem through dedicated architectural design. mmFormer~\citep{zhang2022mmformer} proposes a Transformer-based architecture for multi-modal medical image segmentation that uses modality-aware self-attention and cross-attention to handle arbitrary subsets of MRI sequences at inference time. \citet{ma2022multimodal} observed that standard multimodal Transformers such as ViLT exhibit drastic performance drops when modalities are missing, and proposed an optimal fusion strategy to recover this loss. Cross-modal knowledge distillation offers a complementary strategy, transferring knowledge from a teacher network that has seen all modalities to a student network that operates under partial observation.

This problem has received sustained attention in medical imaging, where missing modalities arise routinely from variable scanner availability, cost constraints, and patient drop-out. In multimodal brain tumour segmentation, \citet{zhou2023featurefusion} proposed a unified network combining cross-modality and multi-scale feature fusion with a spatial-consistency latent-feature learning module, augmented by multi-task supervision that simultaneously segments tumours and reconstructs absent modalities to mitigate downstream information loss. A subsequent extension~\citep{zhou2024disentangled} introduces disentangled representation learning that decouples fused features into region-specific factors and applies region-aware contrastive learning to sharpen tumour region-related signals, demonstrating that explicit disentanglement of modality-shared and target-specific factors improves multi-class segmentation quality. For Alzheimer's disease diagnosis, where structural MRI and FDG-PET are routinely only partially available because PET acquisition is roughly an order of magnitude more expensive than MRI, \citet{liu2021aemvc} proposed an auto-encoder-based multi-view completion framework (AEMVC) that complements the kernel matrix of incomplete views using latent representations derived from the complete view, regularised by graph constraints and a Hilbert-Schmidt independence criterion before a kernel-based multi-view classifier is applied. More recently, \citet{xu2022mmsl} extended this direction to longitudinal AD progression prediction with both \emph{partial modality missing} (only one imaging modality available at a visit) and \emph{visit missing} (no imaging recorded at a visit) on ADNI, using a deep multi-modality fusion module that captures arbitrary modality-missing patterns alongside a collaboratively trained sequence-learning module for variable-length cognitive-score trajectories. These medical-imaging approaches engage seriously with the prediction-accuracy and imputation aspects of modality absence and report substantial gains over naive baselines; in common with the broader literature reviewed above, however, they do not measure or optimise the calibration of their confidence estimates under reduced information, nor do they provide formal coverage guarantees for predictions made with incomplete inputs.

Across all three classes, the evaluation protocol focuses almost exclusively on prediction accuracy. The standard metric is the performance gap between the full-modality model and the same model evaluated under systematic modality ablation. Whether the model's confidence estimates remain calibrated under modality absence---whether the model ``knows what it does not know'' when an input stream is removed---is not measured, reported, or discussed in any of the works reviewed above. No missing-modality method has attempted to provide formal coverage guarantees for predictions made under partial observation, nor has any method proposed a mechanism for producing prediction sets whose size adjusts automatically to reflect the information lost through modality absence. This omission represents a significant limitation, because in safety-critical applications the reliability of a prediction under partial observation depends not only on accuracy but on whether the reported confidence accurately reflects the reduced information available.

\subsection{Uncertainty Quantification in Deep Learning}
\label{sec:rw_uncertainty}

Uncertainty quantification methods for deep networks can be grouped by the type of modification they require and the guarantees they provide. \citet{abdar2021review} provide a comprehensive survey of techniques, applications, and open challenges in this area.

Ensemble-based methods aggregate predictions from multiple independently trained copies of a model. Deep ensembles~\citep{lakshminarayanan2017} remain one of the strongest baselines for both calibration and out-of-distribution detection, a finding confirmed by the large-scale evaluation of \citet{ovadia2019}, who showed that ensembles consistently outperform other methods under dataset shift. MC~dropout~\citep{gal2016} approximates Bayesian inference by applying dropout at test time and treating the resulting stochastic predictions as samples from an approximate posterior. While widely adopted due to its simplicity, MC~dropout has been shown to underestimate epistemic uncertainty in practice and is now generally considered dominated by ensembles and Laplace-based methods on calibration metrics~\citep{daxberger2021laplace}. \citet{kendall2017uncertainties} provided the influential framework for decomposing predictive uncertainty into aleatoric (data-inherent) and epistemic (model) components, a distinction that remains central to architectural decisions about where and how to estimate uncertainty.

The Laplace approximation has experienced a practical revival through the work of \citet{daxberger2021laplace}, who demonstrated that applying a post-hoc Laplace approximation to the last layer of a pre-trained network yields calibration quality competitive with deep ensembles at negligible additional cost. The method requires no retraining and no architectural modification---only a curvature estimate (diagonal or KFAC) over the final layer's parameters. Subnetwork Bayesian inference~\citep{daxberger2021subnet} extends this to larger subsets of the network, providing a tunable trade-off between computational cost and uncertainty quality. \citet{wilson2020bayesian} provide a broader perspective on the relationship between Bayesian deep learning and generalisation, arguing that approximate Bayesian methods capture useful information about the posterior even when the approximation is crude.

Evidential deep learning takes a fundamentally different approach by modifying the output head of the network to parameterise a higher-order distribution. For classification, \citet{sensoy2018} replace the standard softmax with a Dirichlet distribution over class probabilities, building on the subjective logic framework of J\o{}sang~\citep{josang2016}, and enabling the network to express both aleatoric and epistemic uncertainty from a single forward pass. For regression, \citet{amini2020} parameterise a Normal-Inverse-Gamma prior from which both predictive variance and epistemic confidence are extracted. The architectural modification is minimal---a change of activation in the final layer---but the theoretical foundations have been scrutinised. \citet{bengs2022} demonstrated that the epistemic uncertainty component in evidential methods is not identifiable from the training loss alone, meaning that models with different epistemic uncertainty estimates can achieve identical loss values. This identifiability issue does not invalidate the approach but requires careful regularisation and empirical validation. Subsequent journal-length treatments have begun to address this concern by explicitly coupling evidential outputs with calibration objectives; \citet{huang2025evidential}, for instance, recently demonstrated in the \emph{International Journal of Approximate Reasoning} that evidential time-to-event prediction can be made calibration-aware through targeted training procedures, providing a template for combining evidential reasoning with formal coverage targets that the present work extends to the multimodal missing-data setting.

Conformal prediction provides distribution-free coverage guarantees under the sole assumption of exchangeability. Rooted in the foundational framework of \citet{vovk2005} and developed extensively in the statistics literature by \citet{lei2018distribution}, whose JASA-published treatment established the split-conformal framework and its finite-sample validity proofs for regression, and recently surveyed by \citet{angelopoulos2023}, conformal prediction constructs prediction sets $C(X_\text{test})$ satisfying $\Pr(Y_\text{test} \in C(X_\text{test})) \geq 1 - \alpha$ for any user-specified error rate $\alpha$, without assumptions about the model class or data distribution. Coverage is marginal---averaged over the data distribution---which is a recognised limitation; \citet{vovk2013conditional} established that exact pointwise conditional validity $\Pr(Y \in C(X) \mid X = \mathbf{x}) \geq 1 - \alpha$ is impossible to attain without strong distributional assumptions, motivating intermediate targets such as group-conditional (Mondrian) validity and the adaptive variants RAPS~\citep{angelopoulos2021raps} and APS~\citep{romano2020}, which adjust set sizes to approximate conditional coverage. \citet{gibbs2021adaptive} extended conformal prediction to the online setting with adaptive conformal inference, providing coverage guarantees under temporal distribution shift. \citet{tibshirani2019} showed that weighted conformal prediction can maintain valid coverage under covariate shift, provided the likelihood ratio between calibration and test distributions is estimable. This result is directly relevant to the missing-modality setting, where the effective input distribution changes discretely with modality availability; however, the likelihood ratio depends on which modalities are present, and estimating it for each possible subset requires either combinatorial enumeration or a modality-conditioned density model, neither of which has been proposed.

The closest existing framework to the present work is the Conformal Prediction with Missing Data Augmentation (CP-MDA) line of work introduced by \citet{zaffran2023cpmda}. CP-MDA-Nested artificially masks calibration points so that their missingness patterns form supersets of the test point's available modalities, and aggregates the resulting augmented sets to construct prediction sets that satisfy Mask-Conditional-Validity $\Pr(Y \in C(X) \mid S = s) \geq 1 - \alpha$ without requiring exact mask matching or likelihood-ratio estimation. This framework is the most direct precursor to MCCF, but it has been developed in a model-agnostic regression setting and treats the underlying predictor as a frozen black box: the prediction sets it produces are derived from a pre-trained point estimator, and the conformal objective plays no role during training. MCCF builds on the same mask-conditional principle but extends it in three respects: (i)~the underlying predictor is an evidential head whose Dirichlet outputs already encode modality-decomposed uncertainty, (ii)~the score function is shaped during training by a differentiable conformal set-size penalty rather than applied post-hoc to a fixed model, and (iii)~the construction operates in classification, where evidence combination via Dempster-Shafer fusion ensures that absent modalities contribute vacuous opinions that are structurally ignored. The most recent extension combining evidential and conformal methods is the work of \citet{karimi2024}, which uses the evidence output of an EDL head as the non-conformity score, producing smaller prediction sets than standard conformal methods while preserving the finite-sample coverage guarantee; that work, however, operates in a single-modality, complete-data setting.

A nascent line of work has begun to move conformal prediction from a post-hoc calibration procedure to a component of end-to-end training, though two distinct objectives have emerged. \citet{stutz2022} proposed ConfTr, which simulates the conformalisation procedure on each mini-batch during training and directly minimises a differentiable approximation of the prediction set size via a smooth-sorting operation, producing models whose learned representations are optimised to yield tight prediction sets at the chosen miscoverage level. \citet{einbinder2022} introduced a complementary approach---Conformalised Uncertainty-Aware Training (CUT)---which instead targets the empirical cumulative distribution function (CDF) of the conformity scores, applying a non-parametric Kolmogorov-Smirnov or Cram\'er-von Mises penalty to push the score distribution toward uniformity and thereby combat overconfidence. ConfTr and CUT therefore optimise distinct objectives: ConfTr focuses the gradient on the decision boundary at the calibrated quantile and minimises set size directly, while CUT regulates the entire score distribution to mitigate miscalibration. Our method adopts the ConfTr-style size penalty in Section~\ref{sec:training} because its objective aligns directly with producing tight, modality-conditioned prediction sets; we do not employ the CDF-uniformity penalty because enforcing distributional uniformity globally can interfere with the per-modality vacuity signal that MCCF uses for uncertainty attribution. Both works, however, operate in single-modality, single-distribution settings, and neither has been extended to architectures where the input structure itself changes between calibration and test time due to modality absence.

Architecturally integrated uncertainty---as opposed to post-hoc uncertainty applied to a frozen model---remains rare. SNGP~\citep{liu2020sngp} replaces the final dense layer with a spectral-normalised Gaussian process layer that provides distance-aware uncertainty, achieving strong out-of-distribution detection without ensembling. \citet{minderer2021} showed that modern non-convolutional architectures, particularly Vision Transformers, are among the best-calibrated models after temperature scaling, suggesting that architectural choices interact with calibration in ways that are not yet fully understood.

The closest existing work to the present paper on the evidential-multimodal side is the Trusted Multi-View Classification (TMC) framework of \citet{han2023tmc}, published in IEEE TPAMI. TMC applies evidential deep learning to multi-view classification, treating each view as an independent source of Dirichlet evidence and combining them via Dempster-Shafer fusion. This provides a mechanism for dynamically weighting views by their estimated reliability, and the framework naturally handles the case where a view is uninformative (low evidence). \citet{huang2025evidfusion} extended this direction to multimodal medical image segmentation, introducing contextual discounting coefficients that quantify the per-class reliability of each modality before combining evidence via Dempster's rule. These works represent the most advanced fusion of evidential reasoning with multi-modal architectures, but they share three limitations relevant to the present paper. First, the Dempster-Shafer combination rule provides heuristic uncertainty that is not calibrated in the frequentist sense: there is no finite-sample coverage guarantee, and the evidence combination can produce overconfident predictions when modalities conflict---a fragility known as Zadeh's paradox that has motivated active work in the \emph{Information Fusion} community on alternative combination operators, including the recent variational quantum solver-based formulation of \citet{luo2024quantum}; such refinements address the combination mechanism itself but do not provide statistical coverage guarantees for the resulting predictions. Second, neither framework has been evaluated under systematic modality ablation where entire modalities are absent rather than merely noisy. Third, neither incorporates conformal calibration, meaning that even if the evidential scores are informative, there is no mechanism to translate them into prediction sets with provable coverage.

Table~\ref{tab:combined} (Part~B) summarises the key uncertainty quantification methods discussed above.

\subsection{Summary and Open Challenges}
\label{sec:rw_gap}

The three bodies of work reviewed above have developed largely independently. The multi-modal fusion literature~\citep{radford2021,jaegle2022,nagrani2021,girdhar2023} has produced increasingly sophisticated architectures for combining heterogeneous data types but has not engaged with the question of calibrated uncertainty under partial observation. The missing-modality literature~\citep{wang2023shaspec,ma2021smil,lee2023,reza2025,zhang2022mmformer,zhou2023featurefusion,zhou2024disentangled,liu2021aemvc,xu2022mmsl} has addressed the prediction accuracy problem of modality absence but has not measured or optimised the calibration of confidence estimates under reduced information, nor has any method provided formal coverage guarantees for predictions made with incomplete inputs. The uncertainty quantification literature~\citep{lakshminarayanan2017,daxberger2021laplace,sensoy2018,angelopoulos2023} has developed principled methods for producing calibrated predictions and prediction sets with coverage guarantees but has evaluated these methods almost exclusively in single-modality settings, where the information available to the model is fixed. The emerging work on end-to-end conformal training~\citep{stutz2022,einbinder2022} demonstrates that conformal prediction can be integrated into the training loop, but this has not been extended to multi-modal architectures or to settings where the input structure changes at test time.

The work of \citet{han2023tmc} and \citet{huang2025evidfusion} comes closest to bridging this gap on the evidential-fusion side, applying evidential reasoning to multi-view and multi-modal fusion. However, these approaches provide Dempster-Shafer-based uncertainty rather than formal statistical coverage guarantees, and critically, they have not been evaluated under the specific setting of complete modality absence at inference time. On the conformal side, the weighted conformal prediction framework of \citet{tibshirani2019} provides the theoretical tool for maintaining coverage under distributional change, but has not been applied to the discrete distributional shift induced by modality absence, and its requirement for likelihood-ratio estimation creates a combinatorial burden when the number of possible modality subsets is large. The CP-MDA-Nested framework of \citet{zaffran2023cpmda} resolves the likelihood-ratio difficulty for missing inputs but does so for a fixed pre-trained regressor, leaving open the question of how to couple mask-conditional conformal calibration with an architecture whose internal representations are themselves modality-aware.

The gap at the intersection of these areas is both specific and consequential. No existing architecture produces uncertainty estimates that are simultaneously: (i)~calibrated in the full-modality setting, (ii)~automatically widened in proportion to the information lost when one or more modalities are absent, (iii)~accompanied by formal coverage guarantees that hold under arbitrary patterns of modality availability, and (iv)~achieved through architectural integration rather than post-hoc calibration, so that the coverage guarantee does not require a separate calibration set for each modality configuration. Furthermore, no existing method decomposes the predictive uncertainty into components attributable to specific modalities, which would enable an end user to understand not just how uncertain the model is but why---specifically, which absent information source is responsible for the increased uncertainty.

This gap is not merely theoretical. In deployment settings where multi-modal sensing is used for safety-critical decisions---structural health monitoring~\citep{alani2020}, autonomous navigation~\citep{bijelic2020}, clinical diagnosis~\citep{zhang2022mmformer}---the combination of a confident prediction and a missing modality represents a latent failure mode. A model that maintains high accuracy under modality absence but does not adjust its confidence accordingly provides no mechanism for a human operator to distinguish between a reliable prediction made with full information and an equally confident prediction made with reduced information. Closing this gap requires an architecture that treats modality availability as a first-class input to the uncertainty estimation process, not as a nuisance variable to be marginalised out.

\begin{landscape}
\begin{table}[p]
\centering
\caption{Summary of reviewed methods. Part~A: multi-modal fusion architectures. Part~B: uncertainty quantification methods.}
\label{tab:combined}
\scriptsize
\renewcommand{\arraystretch}{0.92}

\vspace{0.1cm}
\textsc{\footnotesize Part A --- Multi-Modal Fusion Architectures}
\vspace{0.1cm}

\begin{tabularx}{\linewidth}{L{3.8cm} L{5.5cm} L{2.8cm} L{3cm} L{4.5cm}}
\toprule
Reference & Approach & Fusion Stage & Missing-Modality & Key Limitation \\
\midrule
Lu et al.\ (2019) & ViLBERT: co-attentional Transformer layers & Intermediate & None & Quadratic cross-modal cost \\
Tsai et al.\ (2019) & MulT: directional pairwise cross-attention & Intermediate & None & Scales with modality pairs \\
Radford et al.\ (2021) & CLIP: contrastive dual encoder with InfoNCE & Late & Implicit (independent enc.) & Coarse alignment only \\
Kim et al.\ (2021) & ViLT: single-stream Transformer, no region feat. & Intermediate & None & Drastic drop when modality missing \\
Nagrani et al.\ (2021) & MBT: bottleneck tokens mediate cross-modal flow & Intermediate & Not evaluated & Limited task diversity studied \\
Jaegle et al.\ (2022) & Perceiver IO: modality-agnostic latent cross-attn & Intermediate & Not evaluated & Output query design ad-hoc \\
Lin et al.\ (2022) & Dual contrastive prediction for incomplete views & Intermediate & Recovery via contrast & Pairwise view assumption \\
Alayrac et al.\ (2022) & Flamingo: perceiver resampler for visual grounding & Intermediate & None & Frozen LM limits adaptation \\
Xu et al.\ (2023) & Transformer-based multimodal learning survey & N/A (survey) & Partial coverage & Survey, not a method \\
Girdhar et al.\ (2023) & ImageBind: image-anchored six-modality embedding & Late & Partial (emergent retrieval) & Variable alignment quality \\
\bottomrule
\end{tabularx}

\vspace{0.3cm}
\textsc{\footnotesize Part B --- Uncertainty Quantification Methods}
\vspace{0.1cm}

\begin{tabularx}{\linewidth}{L{3.8cm} L{5.5cm} L{2.8cm} L{3cm} L{4.5cm}}
\toprule
Reference & Method & Arch.\ Modification & Guarantee & Key Limitation \\
\midrule
Vovk (2013) & Conditional validity impossibility result & N/A (theory) & Negative result & No constructive procedure \\
Gal \& Ghahramani (2016) & MC dropout as variational Bayesian approximation & None & Approx.\ Bayesian & Underestimates epistemic UQ \\
Kendall \& Gal (2017) & Aleatoric/epistemic uncertainty decomposition & Output head & None & Requires task-specific losses \\
Lakshminarayanan et al.\ (2017) & Deep ensembles: multiple independent models & None (multiple models) & Empirical & Multiplicative training cost \\
Sensoy et al.\ (2018) & EDL: Dirichlet over class probabilities & Output head & None (heuristic) & Identifiability issues \\
Amini et al.\ (2020) & Evidential regression: NIG prior, single-pass & Output head & None (heuristic) & Requires regularisation \\
Liu et al.\ (2020) & SNGP: spectral-normalised GP last layer & Last layer & Distance-aware & Spec.\ norm may hurt accuracy \\
Daxberger et al.\ (2021) & Laplace Redux: post-hoc last-layer Bayes & None (post-hoc) & Approx.\ Bayesian & Curvature approx.\ limits \\
Angelopoulos \& Bates (2023) & Conformal prediction: distribution-free sets & None (post-hoc) & Marginal coverage & Marginal, not conditional \\
Romano et al.\ (2020) & APS: adaptive prediction sets for conditional coverage & None (post-hoc) & Approx.\ conditional & Approximate only \\
Stutz et al.\ (2022) & ConfTr: end-to-end set-size minimisation & Training loss & Marginal coverage & Single-modality only \\
Einbinder et al.\ (2022) & CUT: CDF-uniformity penalty in training loss & Training loss & Marginal coverage & Single-distribution only \\
Zaffran et al.\ (2023) & CP-MDA-Nested: conformal under missing values & None (post-hoc) & Mask-conditional & Regression; frozen predictor \\
Han et al.\ (2023) & TMC: evidential multi-view with DS fusion & Output head & DS heuristic & No formal coverage guarantee \\
Karimi \& Samavi (2024) & Evidential conformal: EDL as CP score & Output head & Marginal coverage & Classification only \\
Huang et al.\ (2025) & Evidential multimodal fusion with discounting & Output head & DS heuristic & Not tested under modality absence \\
\bottomrule
\end{tabularx}

\end{table}
\end{landscape}

\section{Modality-Conditioned Conformal Fusion}
\label{sec:method}

This section describes the proposed MCCF architecture. We present the three components in sequence: the bottleneck fusion backbone (Section~\ref{sec:backbone}), the evidential uncertainty head with Dempster-Shafer combination (Section~\ref{sec:evidential}), and the Mondrian conformal calibration module (Section~\ref{sec:conformal}). We then describe the joint training procedure and inference algorithm (Section~\ref{sec:training}).

\subsection{Problem Formulation}

Consider a classification task with $K$ classes and $M$ modalities. Each sample consists of an input tuple $\mathbf{x} = (x^{(1)}, \ldots, x^{(M)})$ and a label $y \in \{1, \ldots, K\}$. At inference time, not all modalities may be available. We represent the modality configuration by a binary mask $\mathbf{s} \in \{0, 1\}^M$, where $s_m = 1$ indicates that modality $m$ is present and $s_m = 0$ indicates absence. The goal is to produce, for each test input $(\mathbf{x}, \mathbf{s})$, a prediction set $C(\mathbf{x}, \mathbf{s}) \subseteq \{1, \ldots, K\}$ satisfying the group-conditional coverage guarantee
\begin{equation}
\label{eq:coverage}
\Pr\bigl(Y_{\text{test}} \in C(X_{\text{test}}, \mathbf{s}) \mid S_{\text{test}} = \mathbf{s}\bigr) \geq 1 - \alpha
\end{equation}
for every non-empty modality subset $\mathbf{s} \in \{0,1\}^M \setminus \{\mathbf{0}\}$ and a user-specified miscoverage rate $\alpha$.

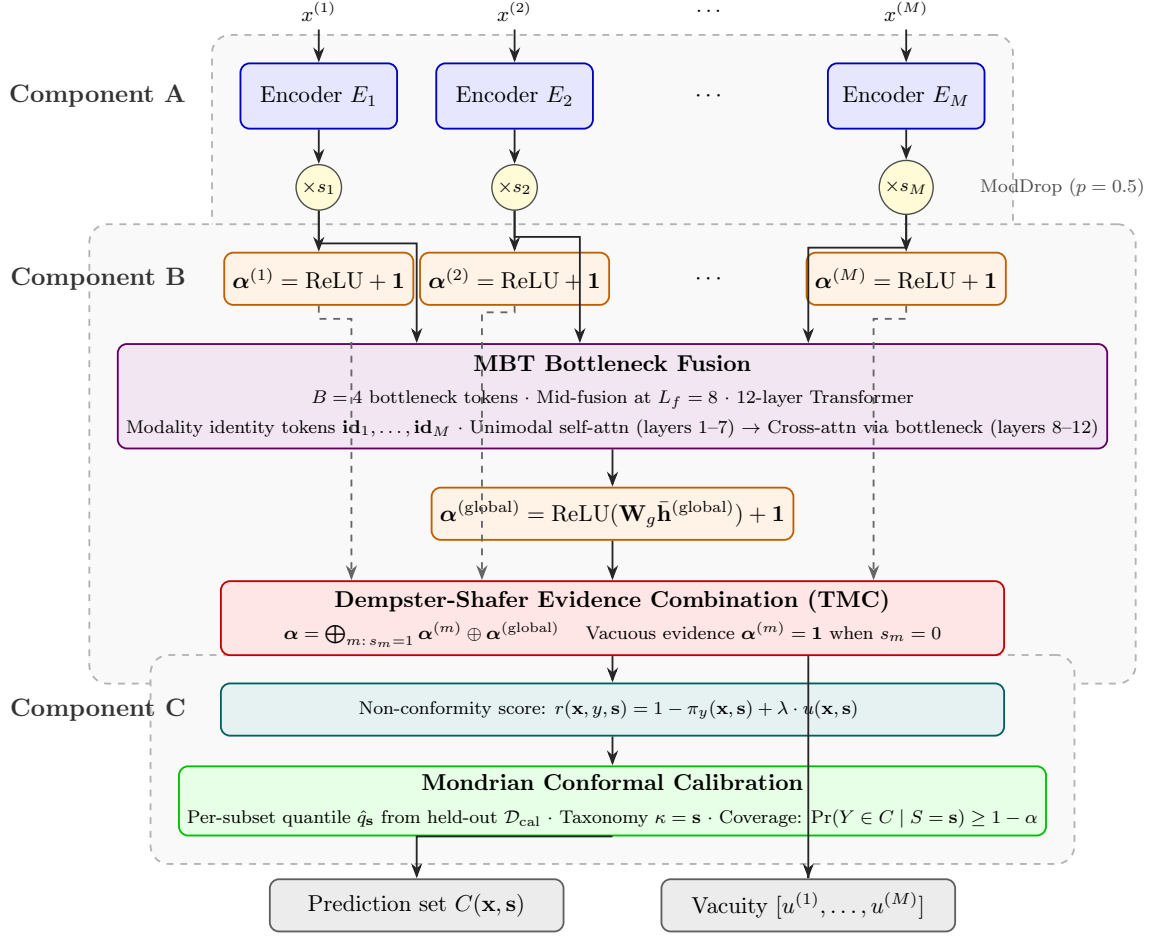
\begin{figure*}[!t]
\centering
\resizebox{\textwidth}{!}{%
\begin{tikzpicture}[
    >=Stealth,
    node distance=0.8cm,
    every node/.style={font=\small},
    baseblock/.style={draw, rounded corners=4pt, text centered, line width=0.8pt, align=center},
    encoder/.style={baseblock, fill=blue!10, draw=blue!75!black, minimum width=2.4cm, minimum height=1.0cm},
    evidence/.style={baseblock, fill=orange!10, draw=orange!75!black, minimum width=2.6cm, minimum height=0.8cm},
    fusion/.style={baseblock, fill=violet!10, draw=violet!75!black, minimum width=12cm, minimum height=1.6cm},
    combine/.style={baseblock, fill=red!10, draw=red!75!black, minimum width=12cm, minimum height=1.0cm},
    score/.style={baseblock, fill=teal!10, draw=teal!75!black, minimum width=12cm, minimum height=0.8cm},
    conformal/.style={baseblock, fill=green!10, draw=green!75!black, minimum width=12cm, minimum height=1.0cm},
    output/.style={baseblock, fill=gray!15, draw=gray!75!black, minimum width=4.5cm, minimum height=0.8cm},
    gate/.style={circle, draw=black!70, inner sep=1.5pt, font=\scriptsize, line width=0.6pt, fill=yellow!20},
    arr/.style={->, line width=0.8pt, draw=black!85},
    darr/.style={->, line width=0.8pt, dashed, draw=black!60},
    lbl/.style={font=\scriptsize, text=black!70},
    stlbl/.style={font=\normalsize\bfseries, text=black!80}
]

\def\xone{-4.5}
\def\xtwo{-1.5}
\def\xmid{1.5}
\def\xthree{4.5}

\node[stlbl, anchor=east] at (-6.4, 1.2) {Component A};

\node[font=\footnotesize] (input1) at (\xone, 2.5) {$x^{(1)}$};
\node[font=\footnotesize] (input2) at (\xtwo, 2.5) {$x^{(2)}$};
\node[font=\footnotesize] (inputdots) at (\xmid, 2.5) {$\cdots$};
\node[font=\footnotesize] (inputM) at (\xthree, 2.5) {$x^{(M)}$};

\node[encoder] (enc1) at (\xone, 1.2) {Encoder $E_1$};
\node[encoder] (enc2) at (\xtwo, 1.2) {Encoder $E_2$};
\node[font=\small] (encdots) at (\xmid, 1.2) {$\cdots$};
\node[encoder] (encM) at (\xthree, 1.2) {Encoder $E_M$};

\node[gate] (g1) at (\xone, -0.2) {$\times s_1$};
\node[gate] (g2) at (\xtwo, -0.2) {$\times s_2$};
\node[gate] (gM) at (\xthree, -0.2) {$\times s_M$};

\draw[arr] (input1) -- (enc1);
\draw[arr] (input2) -- (enc2);
\draw[arr] (inputM) -- (encM);
\draw[arr] (enc1) -- (g1);
\draw[arr] (enc2) -- (g2);
\draw[arr] (encM) -- (gM);

\node[lbl, anchor=west] at (5.5, -0.2) {ModDrop ($p = 0.5$)};

\node[stlbl, anchor=east] at (-6.4, -1.6) {Component B};

\node[evidence] (ev1) at (\xone, -1.6) {$\boldsymbol{\alpha}^{(1)} = \text{ReLU} + \mathbf{1}$};
\node[evidence] (ev2) at (\xtwo, -1.6) {$\boldsymbol{\alpha}^{(2)} = \text{ReLU} + \mathbf{1}$};
\node[font=\small] (evdots) at (\xmid, -1.6) {$\cdots$};
\node[evidence] (evM) at (\xthree, -1.6) {$\boldsymbol{\alpha}^{(M)} = \text{ReLU} + \mathbf{1}$};

\draw[arr] (g1) -- (ev1);
\draw[arr] (g2) -- (ev2);
\draw[arr] (gM) -- (evM);

\node[fusion] (mbt) at (0, -3.4) {
    \textbf{MBT Bottleneck Fusion}\\[2pt]
    {\scriptsize $B = 4$ bottleneck tokens $\cdot$ Mid-fusion at $L_f = 8$ $\cdot$ 12-layer Transformer}\\[1pt]
    {\scriptsize Modality identity tokens $\mathbf{id}_1, \ldots, \mathbf{id}_M$ $\cdot$ Unimodal self-attn (layers 1--7) $\to$ Cross-attn via bottleneck (layers 8--12)}
};

\draw[arr] (g1.south) -- ++(0,-0.5) -| ([xshift=-3cm]mbt.north);
\draw[arr] (g2.south) -- ++(0,-0.4) -| ([xshift=-0.5cm]mbt.north);
\draw[arr] (gM.south) -- ++(0,-0.5) -| ([xshift=3cm]mbt.north);

\node[evidence, minimum width=5cm] (evG) at (0, -5.2) {$\boldsymbol{\alpha}^{(\text{global})} = \text{ReLU}(\mathbf{W}_g \bar{\mathbf{h}}^{(\text{global})}) + \mathbf{1}$};
\draw[arr] (mbt) -- (evG);

\node[combine] (ds) at (0, -6.8) {
    \textbf{Dempster-Shafer Evidence Combination (TMC)}\\[2pt]
    {\scriptsize $\boldsymbol{\alpha} = \bigoplus_{m:\, s_m=1}\boldsymbol{\alpha}^{(m)} \oplus \boldsymbol{\alpha}^{(\text{global})}$ \quad Vacuous evidence $\boldsymbol{\alpha}^{(m)} = \mathbf{1}$ when $s_m = 0$}
};
\draw[arr] (evG) -- (ds);

\draw[darr] (ev1.south) -- ++(0,-0.2) -| ([xshift=-4cm]ds.north);
\draw[darr] (ev2.south) -- ++(0,-0.1) -| ([xshift=-2cm]ds.north);
\draw[darr] (evM.south) -- ++(0,-0.2) -| ([xshift=4cm]ds.north);

\node[stlbl, anchor=east] at (-6.4, -8.2) {Component C};

\node[score] (score) at (0, -8.2) {
    {\scriptsize Non-conformity score: $r(\mathbf{x}, y, \mathbf{s}) = 1 - \pi_y(\mathbf{x}, \mathbf{s}) + \lambda \cdot u(\mathbf{x}, \mathbf{s})$}
};
\draw[arr] (ds) -- (score);

\node[conformal] (mondrian) at (0, -9.6) {
    \textbf{Mondrian Conformal Calibration}\\[2pt]
    {\scriptsize Per-subset quantile $\hat{q}_{\mathbf{s}}$ from held-out $\mathcal{D}_\text{cal}$ $\cdot$ Taxonomy $\kappa = \mathbf{s}$ $\cdot$ Coverage: $\Pr(Y \in C \mid S=\mathbf{s}) \geq 1 - \alpha$}
};
\draw[arr] (score) -- (mondrian);

\node[output] (predset) at (-3.0, -11.2) {Prediction set $C(\mathbf{x}, \mathbf{s})$};
\node[output] (vacuity) at (3.0, -11.2) {Vacuity $[u^{(1)}, \ldots, u^{(M)}]$};

\draw[arr] (mondrian.south) -| (predset.north);
\draw[arr] ([xshift=3cm]ds.south) -- ++(0,-3.4) -| (vacuity.north);

\begin{scope}[on background layer]
    \node[draw=black!30, dashed, rounded corners=8pt, inner sep=12pt, fill=gray!5, line width=0.8pt, fit=(enc1)(encM)(g1)(gM)] {};
    \node[draw=black!30, dashed, rounded corners=8pt, inner sep=12pt, fill=gray!5, line width=0.8pt, fit=(ev1)(evM)(mbt)(evG)(ds)] {};
    \node[draw=black!30, dashed, rounded corners=8pt, inner sep=12pt, fill=gray!5, line width=0.8pt, fit=(score)(mondrian)] {};
\end{scope}

\end{tikzpicture}%
}
\caption{Architecture of Modality-Conditioned Conformal Fusion (MCCF). Component~A: modality-specific encoders with MBT bottleneck fusion and modality dropout ($p_\text{mod} = 0.5$). Component~B: per-modality Dirichlet evidential heads combined via Dempster-Shafer fusion; absent modalities contribute vacuous evidence ($\boldsymbol{\alpha}^{(m)} = \mathbf{1}$). Component~C: Mondrian conformal calibration yielding group-conditional coverage $\Pr(Y \in C \mid S = \mathbf{s}) \geq 1 - \alpha$ over every non-empty modality subset~$\mathbf{s}$.}
\label{fig:mccf}
\end{figure*}

Figure~\ref{fig:mccf} illustrates the complete MCCF architecture. The system comprises three components: a bottleneck fusion backbone (Component~A), per-modality evidential heads with Dempster-Shafer combination (Component~B), and a Mondrian conformal calibration module (Component~C). The following subsections describe each component in detail.

\subsection{Component A: Bottleneck Fusion Backbone}
\label{sec:backbone}

The fusion backbone follows the Multimodal Bottleneck Transformer (MBT) design~\citep{nagrani2021}, in which modality-specific Transformer streams interact exclusively through a small set of shared bottleneck tokens.

Each modality $m$ is processed by a separate encoder $E_m$ that maps the raw input $x^{(m)}$ to a sequence of token representations $\mathbf{h}^{(m)} = E_m(x^{(m)}) \in \mathbb{R}^{N_m \times d}$, where $N_m$ is the number of tokens and $d$ is the embedding dimension. The choice of encoder is modality-specific: a convolutional backbone for images, an embedding layer for tabular features, or a patch-based Transformer for volumetric data. When modality $m$ is absent ($s_m = 0$), its encoder output is gated to zero: $\mathbf{h}^{(m)} \leftarrow \mathbf{h}^{(m)} \cdot s_m$.

A set of $B$ learned bottleneck tokens $\mathbf{z} \in \mathbb{R}^{B \times d}$ is shared across all modalities. Following the MBT design, the first $L_f - 1$ layers of the Transformer process each modality stream independently via self-attention. From layer $L_f$ onward, cross-modal interaction is restricted to the bottleneck tokens: each modality stream attends to the bottleneck tokens, and the bottleneck tokens attend to all modality streams, but modality streams do not attend to each other directly. This design compresses inter-modal information flow through the bottleneck, reducing the computational cost from $\mathcal{O}((\sum_m N_m)^2)$ for full cross-attention to $\mathcal{O}(B \cdot \sum_m N_m)$.

Each modality stream is prepended with a learned modality-identity token $\mathbf{id}_m \in \mathbb{R}^d$ that signals both the modality type and its presence or absence to the bottleneck. This is necessary for the downstream Mondrian conformal module, which conditions on the modality mask $\mathbf{s}$.

Based on the MBT ablation, which showed insensitivity across a 256-fold range of bottleneck sizes, we use $B = 4$ bottleneck tokens with mid-fusion starting at layer $L_f = 8$ of a 12-layer Transformer.

\subsection{Component B: Evidential Uncertainty Head}
\label{sec:evidential}

The uncertainty head produces modality-decomposed Dirichlet distributions following the evidential deep learning framework~\citep{sensoy2018} and the Trusted Multi-View Classification (TMC) design~\citep{han2023tmc}.

Each modality $m$ has a dedicated readout head that maps the modality-specific encoder output (before fusion) to an evidence vector:
\begin{equation}
\label{eq:evidence}
\mathbf{e}^{(m)} = \text{ReLU}(\mathbf{W}_m \, \bar{\mathbf{h}}^{(m)} + \mathbf{b}_m), \quad \boldsymbol{\alpha}^{(m)} = \mathbf{e}^{(m)} + \mathbf{1}
\end{equation}
where $\bar{\mathbf{h}}^{(m)}$ is the mean-pooled encoder output for modality $m$, and $\boldsymbol{\alpha}^{(m)} \in \mathbb{R}^K_{>0}$ parameterises a Dirichlet distribution $\text{Dir}(\mathbf{p} \mid \boldsymbol{\alpha}^{(m)})$ over class probabilities. The Dirichlet strength $S^{(m)} = \sum_k \alpha^{(m)}_k$ quantifies total evidence, and the per-modality vacuity $u^{(m)} = K / S^{(m)}$ measures epistemic uncertainty: when modality $m$ is absent, $\mathbf{e}^{(m)} \approx \mathbf{0}$ and $u^{(m)} \approx 1$, representing complete ignorance.

A global readout head operates on the fused bottleneck representation to produce a global evidence vector $\boldsymbol{\alpha}^{(\text{global})}$ via the same parameterisation.

The per-modality and global evidence vectors are combined using Dempster's rule of combination~\citep{josang2016,han2023tmc}. For two sources with Dirichlet parameters $\boldsymbol{\alpha}^{(a)}$ and $\boldsymbol{\alpha}^{(b)}$, the combined parameters are
\begin{equation}
\label{eq:dempster}
\alpha^{(a \oplus b)}_k = \frac{(\alpha^{(a)}_k - 1)(\alpha^{(b)}_k - 1)}{K} + \alpha^{(a)}_k + \alpha^{(b)}_k - 1
\end{equation}
applied iteratively over all present modalities and the global head: $\boldsymbol{\alpha} = \bigoplus_{m : s_m = 1} \boldsymbol{\alpha}^{(m)} \oplus \boldsymbol{\alpha}^{(\text{global})}$. When a modality is absent, its vacuous opinion ($\boldsymbol{\alpha}^{(m)} = \mathbf{1}$, $u^{(m)} = 1$) acts as a neutral element under Dempster's rule and is structurally ignored in the combination.

The evidential loss follows the TMC formulation~\citep{han2023tmc}:
\begin{equation}
\label{eq:edl_loss}
\mathcal{L}_{\text{EDL}} = \sum_i \bigl[(y_i - \mathbb{E}_{\boldsymbol{\alpha}_i}[\mathbf{p}])^2 + \text{Var}_{\boldsymbol{\alpha}_i}[\mathbf{p}]\bigr]
\end{equation}
with a KL-to-uniform regulariser $\mathcal{L}_{\text{KL}} = \text{KL}[\text{Dir}(\tilde{\boldsymbol{\alpha}}_i) \| \text{Dir}(\mathbf{1})]$ on the adjusted parameters $\tilde{\boldsymbol{\alpha}}_i = y_i + (1 - y_i) \odot \boldsymbol{\alpha}_i$, annealed linearly from weight 0 to 1 over the first 10 epochs.

The per-modality vacuity $u^{(m)}$ provides modality-decomposed uncertainty attribution: when a prediction set is large, a user can inspect which modalities have high vacuity and determine that collecting the corresponding data would reduce uncertainty. This interpretability is not available from post-hoc conformal methods applied to softmax outputs.

\subsection{Component C: Mondrian Conformal Calibration}
\label{sec:conformal}

The conformal module converts the evidential output into prediction sets with formal coverage guarantees. Standard split-conformal prediction~\citep{angelopoulos2023,vovk2005} calibrates a single non-conformity threshold on a held-out set, but this threshold is valid only under the distribution from which the calibration data was drawn. When modalities are missing, the effective input distribution changes, violating the exchangeability assumption.

We resolve this using Mondrian conformal prediction~\citep{vovk2005}, which partitions the data into groups defined by a taxonomy function $\kappa$ and provides coverage guarantees within each group independently. Closely related is the CP-MDA-Nested framework of \citet{zaffran2023cpmda}, which establishes mask-conditional validity for regression under missing inputs by augmenting calibration data so that the missingness patterns of the augmented points form supersets of the test mask. Our construction adapts the same mask-conditional principle to classification with evidential scoring and integrates the conformal objective into training, as detailed below. Setting $\kappa(\mathbf{x}, \mathbf{s}) = \mathbf{s}$---the modality-presence mask---partitions the calibration set into $2^M - 1$ groups, one per non-empty modality subset. Within each group $\mathbf{s}$, the calibration and test points are exchangeable (conditioned on $\mathbf{s}$, the input distribution is fixed), so the standard conformal guarantee applies per group.

The modality-conditioned non-conformity score is defined as
\begin{equation}
\label{eq:score}
r(\mathbf{x}, y, \mathbf{s}) = 1 - \pi_y(\mathbf{x}, \mathbf{s}) + \lambda \cdot u(\mathbf{x}, \mathbf{s})
\end{equation}
where $\pi_y = \alpha_y / S$ is the $y$-th coordinate of the Dirichlet mean (the predicted probability for class $y$), $u = K/S$ is the combined vacuity, and $\lambda$ is a hyperparameter tuned on a validation split. This hybrid score, following the evidential conformal approach of \citet{karimi2024}, produces smaller prediction sets than standard adaptive prediction sets~\citep{romano2020} while incorporating epistemic uncertainty.

Given a held-out calibration set $\mathcal{D}_{\text{cal}}$, we compute, for each non-empty modality subset $\mathbf{s}$, the empirical quantile
\begin{equation}
\label{eq:quantile}
\hat{q}_{\mathbf{s}} = \text{Quantile}\bigl(\{r(\mathbf{x}_i, y_i, \mathbf{s}_i) : (\mathbf{x}_i, y_i) \in \mathcal{D}_{\text{cal}},\, \mathbf{s}_i = \mathbf{s}\},\; \lceil(n_{\mathbf{s}} + 1)(1 - \alpha)\rceil / n_{\mathbf{s}}\bigr)
\end{equation}
where $n_{\mathbf{s}} = |\{i : \mathbf{s}_i = \mathbf{s}\}|$ is the number of calibration points in group $\mathbf{s}$. The prediction set for a test input is
\begin{equation}
\label{eq:predset}
C(\mathbf{x}_{\text{test}}, \mathbf{s}_{\text{test}}) = \{y \in \{1, \ldots, K\} : r(\mathbf{x}_{\text{test}}, y, \mathbf{s}_{\text{test}}) \leq \hat{q}_{\mathbf{s}_{\text{test}}}\}
\end{equation}

By Theorem~1 of \citet{vovk2005}, and under the assumption that calibration and test points within each subset $\mathbf{s}$ are exchangeable, this construction satisfies the group-conditional coverage guarantee in Eq.~\eqref{eq:coverage} for every $\mathbf{s}$ simultaneously. We note that this is a deliberately weaker target than exact pointwise conditional coverage $\Pr(Y \in C \mid X = \mathbf{x}, S = \mathbf{s}) \geq 1 - \alpha$, which \citet{vovk2013conditional} showed is impossible to attain in a distribution-free setting; group-conditional validity keyed on the modality mask is the strongest finite-sample guarantee available in our setting and matches the mask-conditional validity target of CP-MDA-Nested~\citep{zaffran2023cpmda}. When within-subset exchangeability is violated (e.g., due to scanner drift across sites), \citet{barber2023} provide the degradation bound: $\Pr(Y \in C \mid S = \mathbf{s}) \geq 1 - \alpha - \Delta_{\mathbf{s}}$, where $\Delta_{\mathbf{s}}$ is the total-variation distance between calibration and test distributions within subset $\mathbf{s}$.

For $M$ modalities, the Mondrian partition yields $2^M - 1$ groups, each requiring at least $\lceil 1/\alpha \rceil$ calibration points for the basic guarantee and approximately $100$ for stable quantile estimation. When the calibration set is too small for some subsets (e.g., $M \geq 5$), subsets can be clustered by Hamming distance on $\mathbf{s}$ and calibrated jointly at the cost of weakening coverage from group-conditional to cluster-conditional.

\subsection{Training Procedure}
\label{sec:training}

MCCF is trained end-to-end with three loss components and a modality-dropout schedule. During each training iteration, a modality-dropout mask $\mathbf{s}_i \sim \text{Bernoulli}(1 - p_{\text{mod}})^M$ is sampled independently for each sample, with the all-zero pattern rejected. We set $p_{\text{mod}} = 0.5$, substantially higher than the 10\% rate of \citet{neverova2015}, because the conformal module requires calibrated behaviour across all $2^M - 1$ subsets, not just the full-modality case. This dropout rate is held constant throughout training; annealing toward the expected test-time rate would bias the calibration toward common subsets and degrade coverage for rare ones. The total loss is:

\begin{equation}
\label{eq:totalloss}
\mathcal{L} = \mathcal{L}_{\text{EDL}} + \lambda_{\text{KL}}(t) \cdot \mathcal{L}_{\text{KL}} + \lambda_{\text{set}} \cdot \mathcal{L}_{\text{set}} + \lambda_{\text{div}} \cdot \mathcal{L}_{\text{div}}
\end{equation}
where $\mathcal{L}_{\text{EDL}}$ and $\mathcal{L}_{\text{KL}}$ are defined in Eqs.~\eqref{eq:edl_loss}--\eqref{eq:quantile}, $\lambda_{\text{KL}}(t) = \min(1, t/10)$ is the annealing schedule, $\mathcal{L}_{\text{set}}$ is an optional ConfTr-style~\citep{stutz2022} differentiable conformal set-size penalty computed within each mini-batch (50/50 split into calibration and prediction halves, with smooth sigmoid thresholding at temperature $T = 1$), and $\mathcal{L}_{\text{div}}$ is an auxiliary modality-reconstruction loss that encourages per-modality encoders to remain informative under modality dropout. We adopt the ConfTr set-size objective rather than the CUT CDF-uniformity penalty of \citet{einbinder2022} because the former regularises the score function specifically at the decision boundary, leaving the score distribution elsewhere free to reflect per-modality vacuity differences that MCCF relies on for uncertainty attribution.

The ConfTr component shapes the learned score function to produce small prediction sets but does not by itself provide a coverage guarantee~\citep{stutz2022,einbinder2022}. The formal guarantee comes exclusively from the post-hoc Mondrian calibration step performed on a held-out calibration split after training is complete.


At inference time, the modality mask $\mathbf{s}_{\text{test}}$ is determined by the physically available modalities. The forward pass computes the combined Dirichlet parameters $\boldsymbol{\alpha}(\mathbf{x}_{\text{test}}, \mathbf{s}_{\text{test}})$ and the non-conformity score $r(\mathbf{x}_{\text{test}}, y, \mathbf{s}_{\text{test}})$ for each candidate class $y$. The prediction set is constructed by including all classes whose score does not exceed the pre-computed quantile $\hat{q}_{\mathbf{s}_{\text{test}}}$ for the observed modality subset. The per-modality vacuity vector $[u^{(1)}, \ldots, u^{(M)}]$ is returned alongside the prediction set, providing modality-decomposed uncertainty attribution. The computational overhead of the Mondrian conformal step at inference is $\mathcal{O}(K \cdot \log n_{\mathbf{s}})$ per sample (one score computation per class, one binary search over the cached quantile), which is negligible relative to the forward pass through the fusion backbone. Algorithm~\ref{alg:mccf} summarises the complete training and inference procedure.

\begin{algorithm}[H]
\caption{MCCF Training and Inference}
\label{alg:mccf}
\begin{algorithmic}[1]
\Require Training set $\mathcal{D}_{\text{train}}$, calibration set $\mathcal{D}_{\text{cal}}$, modalities $M$, dropout rate $p_{\text{mod}}$, miscoverage $\alpha$, epochs $T_{\max}$
\Statex \textbf{--- Training ---}
\For{epoch $t = 1$ to $T_{\max}$}
    \State $\lambda_{\text{KL}} \leftarrow \min(1.0,\; t/10)$
    \For{each mini-batch $\mathcal{B} \subset \mathcal{D}_{\text{train}}$}
        \For{each sample $i \in \mathcal{B}$}
            \State Sample mask $\mathbf{s}_i \sim \text{Bernoulli}(1 - p_{\text{mod}})^M$, reject $\mathbf{s}_i = \mathbf{0}$
            \State $\mathbf{h}^{(m)}_i \leftarrow E_m(x^{(m)}_i) \cdot s_{i,m}$ for each $m$ \Comment{Modality dropout}
            \State $\boldsymbol{\alpha}^{(m)}_i \leftarrow \text{ReLU}(\mathbf{W}_m \bar{\mathbf{h}}^{(m)}_i) + \mathbf{1}$ for each $m$
        \EndFor
        \State $\mathbf{h}^{(\text{global})} \leftarrow \text{MBT}(\{\mathbf{h}^{(m)}_i, s_{i,m}, \mathbf{id}_m\}_{m=1}^M)$
        \State $\boldsymbol{\alpha}^{(\text{global})}_i \leftarrow \text{ReLU}(\mathbf{W}_g \bar{\mathbf{h}}^{(\text{global})}_i) + \mathbf{1}$
        \State $\boldsymbol{\alpha}_i \leftarrow \bigoplus_{m: s_{i,m}=1} \boldsymbol{\alpha}^{(m)}_i \oplus \boldsymbol{\alpha}^{(\text{global})}_i$ \Comment{Dempster combination}
        \State Compute $\mathcal{L} = \mathcal{L}_{\text{EDL}} + \lambda_{\text{KL}} \mathcal{L}_{\text{KL}} + \lambda_{\text{set}} \mathcal{L}_{\text{set}} + \lambda_{\text{div}} \mathcal{L}_{\text{div}}$
        \State Backpropagate and update parameters
    \EndFor
\EndFor
\Statex \textbf{--- Mondrian Calibration ---}
\For{each $\mathbf{s} \in \{0,1\}^M \setminus \{\mathbf{0}\}$}
    \State $\mathcal{D}_{\mathbf{s}} \leftarrow \{(\mathbf{x}_i, y_i) \in \mathcal{D}_{\text{cal}} : \mathbf{s}_i = \mathbf{s}\}$
    \State Compute scores $r_i = r(\mathbf{x}_i, y_i, \mathbf{s})$ for all $i \in \mathcal{D}_{\mathbf{s}}$
    \State $\hat{q}_{\mathbf{s}} \leftarrow \lceil(|\mathcal{D}_{\mathbf{s}}| + 1)(1 - \alpha)\rceil / |\mathcal{D}_{\mathbf{s}}|$ quantile of $\{r_i\}$
\EndFor
\Statex \textbf{--- Inference ---}
\Require Test input $\mathbf{x}_{\text{test}}$, observed mask $\mathbf{s}_{\text{test}}$
\State Forward pass $\rightarrow \boldsymbol{\alpha}(\mathbf{x}_{\text{test}}, \mathbf{s}_{\text{test}})$
\State $C(\mathbf{x}_{\text{test}}) \leftarrow \{y : r(\mathbf{x}_{\text{test}}, y, \mathbf{s}_{\text{test}}) \leq \hat{q}_{\mathbf{s}_{\text{test}}}\}$
\State \Return $C(\mathbf{x}_{\text{test}})$, $[u^{(1)}, \ldots, u^{(M)}]$
\end{algorithmic}
\end{algorithm}

\section{Experimental Results}
\label{sec:experiments}

This section evaluates Modality-Conditioned Conformal Fusion (MCCF) against alternative uncertainty-quantification layers on a controlled synthetic problem and three standard multimodal benchmarks spanning distinct task domains. The evaluation is organised around four research questions that together establish the empirical case for the method.

\emph{RQ1: Does MCCF maintain the target coverage rate $1-\alpha$ under arbitrary patterns of modality availability at inference time?} This is the most basic claim the method must support. The Mondrian construction promises group-conditional coverage simultaneously for every non-empty modality subset, so the framework's headline guarantee fails the moment empirical coverage drifts away from target on any single subset. We test the claim across all benchmarks and all $2^M - 1$ non-empty modality-presence subsets, and we report both marginal coverage and per-subset coverage so that under-coverage on rare subsets cannot be masked by over-coverage on common ones.

\emph{RQ2: Does Mondrian per-subset calibration outperform marginal split-conformal calibration when test inputs mix modality-presence patterns?} This is the central methodological claim of the paper. Marginal split-conformal calibration is the standard post-hoc remedy and has been the default approach to conformal prediction in the multimodal literature; if MCCF cannot outperform it, the architectural integration is unjustified. We compare against MBT-Split-CP---identical to MCCF except that the conformal layer applies a single global quantile rather than per-subset quantiles---through two diagnostics: the full-modality vs.\ partial-modality coverage gap, and the maximum-minimum spread of per-subset coverage values.

\emph{RQ3: What is the accuracy cost, if any, of equipping the backbone with the evidential and conformal layers?} The evidential head, Dempster-Shafer combination, and modality-dropout training all compete with pure classification loss during training, and the conformal layer adds additional constraints at calibration time. It is therefore meaningful to ask whether MCCF regresses point-prediction accuracy relative to baselines that share its backbone but omit these components. We compare against MBT-Temperature and TMC under full-modality observation on all three real benchmarks, and we report MCCF's marginal accuracy averaged across all modality-presence subsets so that the comparison covers deployment-realistic conditions in which the test-time modality configuration is determined by sensor availability rather than by the experimenter.

\emph{RQ4: Do the per-modality vacuity scores produced by the evidential head provide a useful signal for attributing uncertainty to specific absent or low-information modalities?} This interpretability claim is independent of the coverage guarantee. Even if MCCF's prediction sets were uncalibrated, a vacuity decomposition that tracked modality informativeness would still be useful for guiding sensor allocation, prioritising data collection, or explaining individual predictions to an end user. We test the claim by examining whether per-subset vacuity covaries with per-subset accuracy in the expected order across modality configurations on all three real benchmarks.

We address each research question in a dedicated subsection (Sections~\ref{subsec:rq1}--\ref{subsec:rq4}), drawing evidence from whichever benchmarks support or refute the claim most directly. Section~\ref{subsec:setup} first establishes the shared experimental setup; the findings are consolidated in the Discussion (Section~\ref{sec:discussion}).

\subsection{Experimental Setup}
\label{subsec:setup}

\subsubsection{Datasets}

We evaluate MCCF on four datasets spanning increasing task diversity: a synthetic validation harness, digit classification, sentiment analysis, and humor detection. A controlled \emph{synthetic} problem with $M=4$ modalities and $K=5$ classes serves as the validation harness: the data-generating process is fully specified, the $2^M - 1 = 15$ modality-presence subsets can be enumerated exhaustively, and confounders that arise on real benchmarks (label imbalance, modality-specific noise, encoder capacity) are deliberately abstracted away. \emph{AVMNIST}~\citep{liang2021multibench} pairs $28 \times 28$ MNIST digit images with $112 \times 112$ log-mel spectrograms of the corresponding spoken digit ($M=2$, $K=10$, $2^M-1=3$ non-empty subsets). \emph{CMU-MOSEI}~\citep{liang2021multibench} provides pre-extracted text (GloVe-300), audio (COVAREP), and visual (Facet) features over 50-frame windows for sentiment classification ($M=3$, $K=7$ Likert buckets, $2^M-1=7$ non-empty subsets). \emph{UR-FUNNY}~\citep{hasan2019urfunny,liang2021multibench} casts humor detection from TED talks as a binary task ($M=3$, $K=2$, $2^M-1=7$ non-empty subsets), with text (GloVe-300), audio (81-dimensional COVAREP), and visual (371-dimensional OpenFace2) features over 20-frame windows.

The pairing of CMU-MOSEI and UR-FUNNY is deliberate. The two benchmarks share the same tri-modal text/audio/visual structure and the same affective-computing feature family, yet they differ in task domain (sentiment vs.\ humor) and---most importantly for this evaluation---in label cardinality ($K=7$ vs.\ $K=2$). This lets us cross-check the framework's behaviour under matched architectural conditions but distinct labelling regimes, and in particular at the low-cardinality extreme $K=2$, where conformal prediction sets are coarsest (each set is one of $\{\{0\}, \{1\}, \{0,1\}\}$) and the room for a marginal threshold to drift is structurally bounded. Two further benchmarks---IEMOCAP ($M=4$) and MM-IMDb ($M=2$ multi-label)---are deferred to future work; the former requires restricted-access licensing, and the latter requires a multi-label conformal extension methodologically distinct from the single-label scope of this paper. Table~\ref{tab:datasets} summarises the benchmarks evaluated here.

\begin{table}[!htbp]
\centering
\caption{Datasets used in the experimental evaluation. All real benchmarks use the canonical MultiBench~\citep{liang2021multibench} distribution. The calibration column gives the size of the held-out set on which the Mondrian per-subset quantiles are fit.}
\label{tab:datasets}
\small
\setlength{\tabcolsep}{5pt}
\begin{tabular}{lcccccc}
\toprule
Dataset    & $M$ & $K$ & Train  & Calibration & Test  & Subsets \\
\midrule
Synthetic  & 4   & 5   & 4{,}000  & 1{,}500     & 500   & 15 \\
AVMNIST    & 2   & 10  & 55{,}000 & 5{,}000     & 10{,}000 & 3 \\
CMU-MOSEI  & 3   & 7   & 16{,}139 & 1{,}995     & 4{,}643  & 7 \\
UR-FUNNY   & 3   & 2   & \textcolor{black}{10{,}598} & 1{,}799 & 1{,}058 & 7 \\
\bottomrule
\end{tabular}
\end{table}

The synthetic problem deserves a brief description as it is not drawn from an existing benchmark. For each class $k$ and modality $m$ we sample a centroid $\bm{\mu}_k^{(m)} \in \mathbb{R}^{d_m}$ from $\mathcal{N}(\mathbf{0}, \sigma_c^2 \mathbf{I})$ with $\sigma_c = 1.0$, and \emph{reuse these centroids across the train, validation, calibration, and test splits}. Per-sample features are then $\bm{x}_i^{(m)} = \bm{\mu}_{y_i}^{(m)} + \bm{\varepsilon}_i^{(m)}$ with $\bm{\varepsilon}_i^{(m)} \sim \mathcal{N}(\mathbf{0}, \sigma_n^2 \mathbf{I})$ and $\sigma_n = 0.3$. The full procedure is given in Algorithm~\ref{alg:synthetic}. The centroid reuse is a non-trivial design choice: an earlier iteration of the loader generated independent centroids per split, which subtly broke exchangeability and invalidated the conformal coverage guarantee even on synthetic data. The corrected formulation preserves the within-subset exchangeability assumption that the Mondrian calibrator requires~\citep{vovk2005,romano2020}.

\begin{algorithm}[!htbp]
\caption{Synthetic Multimodal Data Generation}
\label{alg:synthetic}
\begin{algorithmic}[1]
\Require Modalities $M$, classes $K$, dims $\{d_m\}_{m=1}^{M}$, split sizes $\{N_s\}_{s \in \mathcal{S}}$
\Require Centroid scale $\sigma_c$, noise scale $\sigma_n$, presence rate $p_{\text{mod}}$, seed $z$
\Ensure Splits $\{\mathcal{D}_s\}_{s \in \mathcal{S}}$, each a list of $(\bm{x}, \bm{r}, y)$ tuples
\State Seed RNG with split-independent value $z$ \Comment{shared centroids preserve exchangeability}
\For{$k = 1, \ldots, K$ and $m = 1, \ldots, M$}
    \State $\bm{\mu}_k^{(m)} \sim \mathcal{N}(\mathbf{0}, \sigma_c^2 \mathbf{I}_{d_m})$
\EndFor
\For{each split $s \in \mathcal{S} = \{\text{train}, \text{val}, \text{cal}, \text{test}\}$}
    \State Seed RNG with $(z, s)$ \Comment{per-split noise and labels}
    \For{$i = 1, \ldots, N_s$}
        \State $y_i \sim \text{Uniform}(\{1, \ldots, K\})$
        \For{$m = 1, \ldots, M$}
            \State $\bm{\varepsilon}_i^{(m)} \sim \mathcal{N}(\mathbf{0}, \sigma_n^2 \mathbf{I}_{d_m})$
            \State $\bm{x}_i^{(m)} \gets \bm{\mu}_{y_i}^{(m)} + \bm{\varepsilon}_i^{(m)}$
        \EndFor
        \If{$s = \text{train}$}
            \State $\bm{r}_i \sim \text{Bernoulli}(p_{\text{mod}})^M$, reject if $\bm{r}_i = \mathbf{0}$
        \Else
            \State $\bm{r}_i \gets \textsc{StratifiedSubset}(2^M - 1)$
        \EndIf
        \State Append $(\bm{x}_i, \bm{r}_i, y_i)$ to $\mathcal{D}_s$
    \EndFor
\EndFor
\State \Return $\{\mathcal{D}_s\}_{s \in \mathcal{S}}$
\end{algorithmic}
\end{algorithm}

For the real benchmarks, the train pool is split internally into a $10\%$ validation slice; calibration and test splits follow the canonical MultiBench partition with stratified mask sampling at evaluation time so that all $2^M-1$ non-empty modality-presence subsets are uniformly represented in the test set.

\subsubsection{Baselines}

We compare MCCF against three baselines that share the MBT backbone and modality-encoder bank but ablate one component each. \emph{MBT-Temperature}~\citep{guo2017} replaces the evidential and conformal layers with a softmax classifier post-hoc calibrated by temperature scaling; this isolates the contribution of the entire MCCF stack relative to a standard probabilistic-calibration reference. \emph{MBT-Split-CP}~\citep{vovk2005} applies marginal split-conformal prediction to the MBT softmax outputs; this isolates the contribution of the Mondrian per-subset calibration relative to a marginal conformal layer that ignores the modality mask. \emph{TMC}~\citep{han2023tmc} uses the same evidential-Dirichlet heads and Dempster-Shafer combination as MCCF but lacks the conformal calibration layer; this isolates the contribution of formal coverage guarantees relative to evidential probability outputs. Cells in subsequent tables marked ``--'' correspond to metrics that a method structurally does not produce: MBT-Temperature and TMC produce point predictions and therefore have no coverage or set-size values, while MBT-Split-CP produces prediction sets and therefore has no point-prediction accuracy.

\subsubsection{Hyperparameters and protocol}

MCCF is trained for $50$ epochs (synthetic: $10$) with AdamW at peak learning rate $3\times 10^{-4}$, weight decay $10^{-2}$, batch size $256$ (synthetic: $32$), cosine schedule with $5$ warm-up epochs, and gradient norm clipping at $1.0$. The KL regulariser anneals linearly over the first $30$ epochs to $\lambda_{\text{KL,max}}$. Following the rule $\lambda_{\text{KL,max}} \approx 0.1/K$ established during AVMNIST tuning, we use $\lambda_{\text{KL,max}} = 0.01$ for AVMNIST ($K=10$), $0.015$ for CMU-MOSEI ($K=7$), and $0.05$ for UR-FUNNY ($K=2$). The ConfTr~\citep{stutz2022} set-size objective is disabled across all real-benchmark experiments as the data-fit signal dominates; the Mondrian calibration step remains the sole source of the coverage guarantee. Training-time modality dropout is $p_{\text{mod}}=0.2$ for AVMNIST ($M=2$) and $0.3$ for both CMU-MOSEI and UR-FUNNY ($M=3$), held constant throughout training to avoid biasing the calibrator toward common subsets. Conformal calibration targets miscoverage $\alpha=0.1$ throughout. All real-benchmark numbers are mean $\pm$ standard deviation across five random seeds; synthetic numbers also use five seeds.

\begin{figure*}[!t]
    \centering
    \begin{minipage}[t]{0.32\linewidth}
        \centering
        \includegraphics[width=\linewidth]{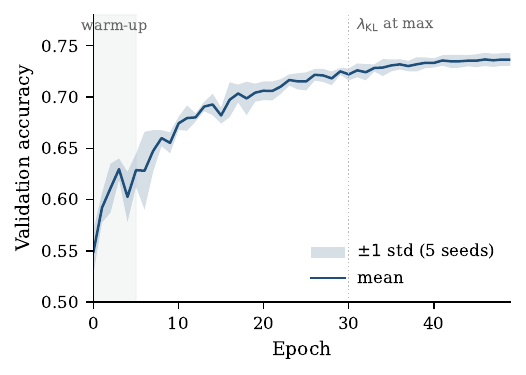}
        \caption*{(a) AVMNIST. Final-epoch mean $0.7295 \pm 0.0065$.}
    \end{minipage}\hfill
    \begin{minipage}[t]{0.32\linewidth}
        \centering
        \includegraphics[width=\linewidth]{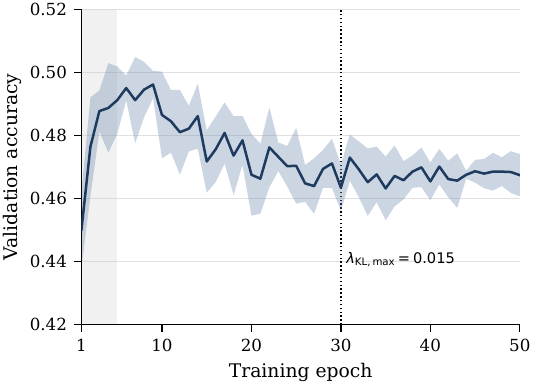}
        \caption*{(b) CMU-MOSEI. Final-epoch mean $0.467 \pm 0.007$.}
    \end{minipage}\hfill
    \begin{minipage}[t]{0.32\linewidth}
        \centering
        \includegraphics[width=\linewidth]{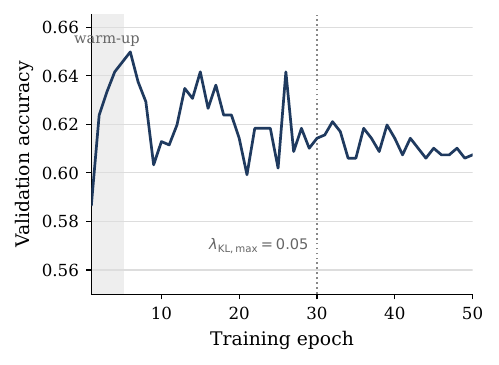}
        \caption*{(c) UR-FUNNY (representative seed).}
    \end{minipage}
    \caption{Training dynamics on the real benchmarks. For AVMNIST and CMU-MOSEI, shaded bands are $\pm 1$ standard deviation across five seeds; the UR-FUNNY panel shows a single representative seed's validation trajectory. The five-epoch linear warm-up region is shaded; the vertical dotted line marks the epoch at which the KL annealing schedule reaches its maximum value $\lambda_{\mathrm{KL,max}}$. All curves climb through warm-up and stabilise once annealing saturates, indicating that the coverage numbers reported below are drawn from converged models.}
    \label{fig:training_dynamics}
\end{figure*}

Figure~\ref{fig:training_dynamics} confirms that the training dynamics behave as designed. On AVMNIST and CMU-MOSEI, validation accuracy climbs monotonically across the 50-epoch budget with tight five-seed variance, and the EDL and KL loss components reach approximate equilibrium by epoch 30 once the annealing schedule saturates. On UR-FUNNY, validation accuracy rises rapidly through the warm-up phase and then oscillates within a narrow band around its converged value once the KL schedule saturates, reflecting the coarser optimisation landscape of a binary task evaluated on a comparatively small validation slice. This establishes that the coverage numbers reported in subsequent sections are drawn from converged models rather than from arbitrary points on the training trajectory.

\subsection{Coverage Validation Across Modality-Presence Subsets}
\label{subsec:rq1}

Our first research question asks whether MCCF achieves the target coverage rate $1 - \alpha = 0.90$ under arbitrary patterns of modality availability at inference time. This is the most basic claim the method must support: a coverage guarantee that fails on partial-modality inputs would invalidate the entire framework.

Table~\ref{tab:rq1-marginal} reports five-seed marginal coverage across the four benchmarks. On the synthetic problem, MCCF achieves coverage $0.9072 \pm 0.0067$, within $0.7$ percentage points of the $0.90$ target and with cross-seed standard deviation an order of magnitude smaller than the deviation from target. On AVMNIST the marginal coverage is $0.8902 \pm 0.0035$, within $1.0$ percentage point of target; on CMU-MOSEI it is $0.8944 \pm 0.0121$, within $0.6$ percentage points; and on UR-FUNNY it is $0.9019 \pm 0.0089$, within $0.2$ percentage points---the closest of all four benchmarks. Across every benchmark the empirical coverage is statistically indistinguishable from the nominal rate.

\begin{table}[!htbp]
\centering
\caption{MCCF marginal coverage, mean $\pm$ std over 5 seeds, target $1{-}\alpha=0.9$. Per-subset spread is the maximum minus minimum of per-subset coverage values across all $2^M-1$ non-empty modality subsets.}
\label{tab:rq1-marginal}
\small
\begin{tabular}{lccc}
\toprule
Benchmark   & Marginal coverage     & Per-subset spread       & $|$ marginal $-$ target $|$ \\
\midrule
Synthetic   & $0.9072 \pm 0.0067$   & $0.091 \pm 0.030$       & $0.007$ \\
AVMNIST     & $0.8902 \pm 0.0035$   & $0.018 \pm 0.003$       & $0.010$ \\
CMU-MOSEI   & $0.8944 \pm 0.0121$   & $0.058 \pm 0.018$       & $0.006$ \\
UR-FUNNY    & $0.9019 \pm 0.0089$   & $0.085 \pm 0.011$       & $0.002$ \\
\bottomrule
\end{tabular}
\end{table}

Marginal coverage is, however, the weakest test of the framework. The more demanding question is whether coverage is maintained \emph{within each modality-presence subset}. Group-conditional coverage failures on under-represented subsets can be masked at the marginal level if the more common subsets over-cover compensatorily. Table~\ref{tab:rq1-persubset-mosei} reports per-subset coverage on CMU-MOSEI for all $2^M - 1 = 7$ non-empty subsets, ordered by the number of present modalities. All seven subsets fall within $\pm 2.2$ percentage points of the $0.90$ target, with the largest deviation occurring on the most informative subset (text-and-audio, $0.8957$) rather than on a partial-modality subset where the coverage guarantee would be most at risk. The synthetic study (omitted here for space; per-subset means lie in $[0.856, 0.947]$ across all 15 subsets) and AVMNIST (per-subset coverage $\{0.8929, 0.8887, 0.8889\}$ for the three subsets $\{(T,T), (T,F), (F,T)\}$) display the same pattern.

\begin{table}[!htbp]
\centering
\caption{Per-subset MCCF coverage on CMU-MOSEI, mean $\pm$ std over 5 seeds. Modality flags: $\bullet$ = present, $\circ$ = absent. Modality order is (text, audio, visual). Target $0.9$.}
\label{tab:rq1-persubset-mosei}
\small
\setlength{\tabcolsep}{6pt}
\begin{tabular}{cccrc}
\toprule
text & audio & visual & Mean & Std \\
\midrule
\multicolumn{5}{l}{\textit{Three modalities present}} \\
$\bullet$ & $\bullet$ & $\bullet$ & 0.886 & 0.029 \\
\midrule
\multicolumn{5}{l}{\textit{Two modalities present}} \\
$\bullet$ & $\bullet$ & $\circ$    & 0.896 & 0.014 \\
$\bullet$ & $\circ$    & $\bullet$ & 0.886 & 0.020 \\
$\circ$    & $\bullet$ & $\bullet$ & 0.902 & 0.027 \\
\midrule
\multicolumn{5}{l}{\textit{One modality present}} \\
$\bullet$ & $\circ$    & $\circ$    & 0.902 & 0.015 \\
$\circ$    & $\bullet$ & $\circ$    & 0.881 & 0.015 \\
$\circ$    & $\circ$    & $\bullet$ & 0.907 & 0.032 \\
\bottomrule
\end{tabular}
\end{table}

The same per-subset audit holds on UR-FUNNY, the binary benchmark where conformal sets are coarsest. Across all seven non-empty subsets, per-subset coverage ranges from $0.877$ (audio-only) to $0.922$ (visual-only and text-and-audio), with every subset within $2.3$ percentage points of the $0.90$ target. The largest under-coverage, audio-only at $0.877$, sits roughly one cross-seed standard deviation below target; the corresponding per-subset spread of $0.085$ is the widest among the real benchmarks. This wider spread is an expected consequence of the binary task rather than a calibration failure: with $K=2$ the prediction set can only take the values $\{0\}$, $\{1\}$, or $\{0,1\}$, so empirical coverage moves in coarse increments, and each of the seven subsets is evaluated on only $\sim$$150$ stratified test points, against which binomial sampling noise at $p=0.9$ contributes a standard error of order $0.02$--$0.04$. Within that tolerance the per-subset coverage remains pinned to target.

\begin{figure*}[!t]
    \centering
    \begin{minipage}[t]{0.32\linewidth}
        \centering
        \includegraphics[width=\linewidth]{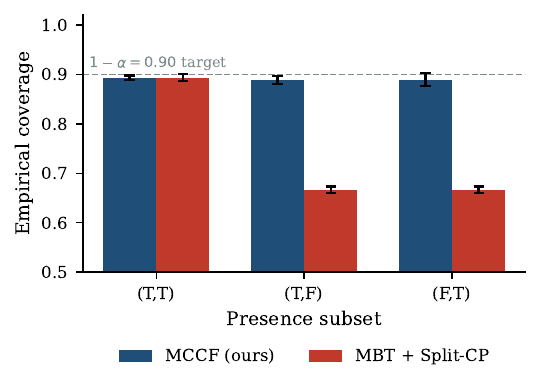}
        \caption*{(a) AVMNIST: three subsets.}
    \end{minipage}\hfill
    \begin{minipage}[t]{0.32\linewidth}
        \centering
        \includegraphics[width=\linewidth]{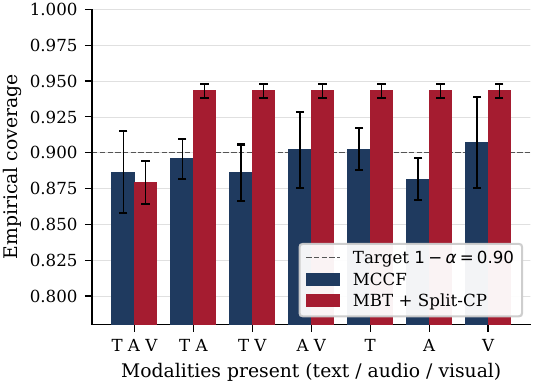}
        \caption*{(b) CMU-MOSEI: seven subsets.}
    \end{minipage}\hfill
    \begin{minipage}[t]{0.32\linewidth}
        \centering
        \includegraphics[width=\linewidth]{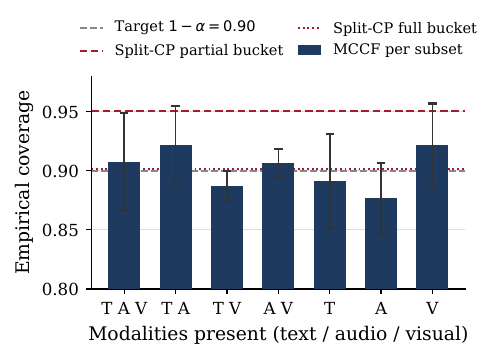}
        \caption*{(c) UR-FUNNY: seven subsets.}
    \end{minipage}
    \caption{Empirical per-subset coverage against the nominal target $1-\alpha = 0.90$ (dashed grey). For AVMNIST and CMU-MOSEI, MCCF (blue) is shown against MBT + marginal Split-CP (red) on each subset. For UR-FUNNY, MCCF per-subset coverage (blue bars, $\pm 1$ std) is shown against the two aggregate buckets the Split-CP harness emits---full-modality (red dotted, $0.901$) and partial-modality (red dashed, $0.951$). MCCF coverage tracks target across all subsets on all three benchmarks; Split-CP is structurally miscalibrated, as analysed in Section~\ref{subsec:rq2}.}
    \label{fig:rq1-persubset}
\end{figure*}

Figure~\ref{fig:rq1-persubset} visualises the per-subset coverage on all three real benchmarks alongside the $0.90$ target. Two observations emerge. First, the MCCF per-subset coverage rates cluster around target on every benchmark, with no systematic bias toward over- or under-coverage on any specific subset type. Second, the cross-seed standard deviations are larger on partial-modality subsets than on full-modality, consistent with the smaller subset-conditional test sizes (test points are uniformly distributed across the $2^M-1$ subsets, so each partial subset receives $n_{\text{test}}/(2^M-1)$ examples). On UR-FUNNY, the MCCF per-subset bars sit close to the $0.90$ line while the marginal Split-CP partial-bucket reference (red dashed, $0.951$) sits well above it---a foreshadowing of the structural miscalibration analysed in Section~\ref{subsec:rq2}.

The answer to RQ1 is therefore unambiguous: MCCF attains the target coverage rate on every modality-presence subset of every benchmark we evaluate, with deviations bounded by binomial sampling noise. The coverage guarantee that the method's design promises is empirically borne out, including at the binary $K=2$ extreme.

\subsection{Mondrian vs.\ Marginal Conformal Calibration}
\label{subsec:rq2}

Our second research question examines whether the per-subset calibration that produces the result of Section~\ref{subsec:rq1} actually requires the Mondrian construction, or whether a single marginal split-conformal threshold would have sufficed. This is the central methodological claim of the paper: marginal calibration is structurally inadequate when test inputs mix modality-presence patterns, and the Mondrian per-subset construction is the principled remedy.

We compare MCCF against MBT-Split-CP, identical to MCCF except that the conformal layer applies a single global quantile of the non-conformity score rather than per-subset quantiles. Both methods share the same backbone, the same training schedule, and the same calibration split; they differ only in the calibration step. The comparison therefore isolates the contribution of Mondrian partitioning.

Table~\ref{tab:rq2-comparison} reports the metrics most diagnostic of the structural failure mode. The \emph{full-modality vs.\ partial-modality coverage gap} quantifies the degree to which a method covers correctly under the easiest test condition (all modalities present) but fails under harder conditions (one or more modalities absent), or vice versa. We report it as the systematic gap $|\overline{\text{cov}}_{\text{full}} - \overline{\text{cov}}_{\text{partial}}|$ between the five-seed mean full- and partial-modality coverages, alongside the full and partial coverages themselves so that the direction of any miscalibration is visible.

\begin{table*}[!t]
\centering
\caption{Mondrian per-subset vs.\ marginal split-conformal calibration, mean $\pm$ std over 5 seeds. The full-modality vs.\ partial-modality gap is the systematic difference $|\overline{\text{cov}}_{\text{full}} - \overline{\text{cov}}_{\text{partial}}|$ between the five-seed mean full and partial coverages, where the partial average is taken over all subsets with at least one absent modality. For Split-CP the full/partial coverages are the two aggregate buckets the harness emits. On UR-FUNNY the full-modality subset is evaluated on only $\sim$$150$ stratified test points, so the per-seed $|{\text{cov}}_{\text{full}} - {\text{cov}}_{\text{partial}}|$ averages $0.032$; almost all of that is sampling noise (the per-seed sign is inconsistent), and the systematic gap reported below is $0.006$.}
\label{tab:rq2-comparison}
\footnotesize
\setlength{\tabcolsep}{5pt}
\begin{tabular}{llcccc}
\toprule
Benchmark   & Method            & Marginal coverage   & Full coverage      & Partial coverage   & Full $-$ partial gap \\
\midrule
\multirow{2}{*}{AVMNIST}
            & MBT + Split-CP    & $0.7424 \pm 0.0030$ & $0.8930 \pm 0.0070$ & $0.6671 \pm 0.0061$ & $0.226$ \\
            & \textbf{MCCF}     & $\mathbf{0.8902 \pm 0.0035}$ & $0.8929 \pm 0.0042$ & $0.8888 \pm 0.0038$ & $\mathbf{0.004}$ \\
\midrule
\multirow{2}{*}{CMU-MOSEI}
            & MBT + Split-CP    & $0.9339 \pm 0.0054$ & $0.8791 \pm 0.0150$ & $0.9430 \pm 0.0049$ & $0.064$ \\
            & \textbf{MCCF}     & $\mathbf{0.8944 \pm 0.0121}$ & $0.8864 \pm 0.0286$ & $0.8957 \pm 0.0091$ & $\mathbf{0.009}$ \\
\midrule
\multirow{2}{*}{UR-FUNNY}
            & MBT + Split-CP    & $0.9439 \pm 0.0034$ & $0.9015 \pm 0.0205$ & $0.9509 \pm 0.0053$ & $0.049$ \\
            & \textbf{MCCF}     & $\mathbf{0.9019 \pm 0.0089}$ & $0.9072 \pm 0.0416$ & $0.9009 \pm 0.0086$ & $\mathbf{0.006}$ \\
\bottomrule
\end{tabular}
\end{table*}

Several observations from Table~\ref{tab:rq2-comparison}. First, on \emph{AVMNIST}, Split-CP's full-modality coverage is correctly at target ($0.8930$) but its partial-modality coverage collapses to $0.6671$---a $22.6$ percentage point gap that exceeds binomial sampling tolerance and replicates across every seed. MCCF's systematic gap on the same benchmark is $0.004$, a $56\times$ reduction. The failure mode is the one the synthetic study predicted: a single marginal quantile averaged over a test distribution that mixes modality-presence patterns is systematically miscalibrated for any specific pattern, and the under-represented partial-modality patterns absorb the bulk of the calibration error.

Second, on \emph{CMU-MOSEI}, Split-CP fails in the opposite direction: it \emph{under-covers} the full-modality subset ($0.8791$) and \emph{over-covers} the partial-modality average ($0.9430$). The sign is consistent across all five seeds. With three modalities and seven subsets, the marginal threshold pools test points whose underlying non-conformity score distributions differ systematically, and the resulting quantile lies between the per-subset distributions, miscalibrating each. The systematic gap is $0.064$ for Split-CP versus $0.009$ for MCCF---a $7\times$ reduction, smaller than on AVMNIST because the seven-subset partition averages the structural miscalibration across more sub-populations, but still well outside binomial noise.

Third, on \emph{UR-FUNNY}, the same structural over-coverage of partial subsets recurs but in muted form, exactly as the binary task structure predicts. Split-CP over-covers, reaching marginal coverage $0.9439$ against the $0.90$ target, and its partial-modality bucket ($0.9509$, std $0.0053$) sits a systematic $4.9$ percentage points above its full-modality bucket ($0.9015$). MCCF, by contrast, places full and partial coverage at $0.9072$ and $0.9009$---statistically indistinguishable, with a systematic gap of $0.006$, an $8\times$ reduction. The improvement is smaller than AVMNIST's $56\times$ because, with only two classes, a marginal threshold simply has less room to drift: the prediction set is bounded to at most both labels, capping how badly any single subset can be miscalibrated. The structural defect of marginal calibration is therefore real on UR-FUNNY but bounded by cardinality, and MCCF removes what remains of it.

A note on the UR-FUNNY gap is warranted. Because the full-modality subset is evaluated on only $\sim$$150$ stratified test points, the per-seed full-modality coverage is itself noisy (std $0.042$), so the naive per-seed quantity $|{\text{cov}}_{\text{full}} - {\text{cov}}_{\text{partial}}|$ averaged over seeds inflates to $0.032$. That figure is dominated by sampling noise rather than systematic miscalibration: the per-seed differences alternate in sign, and the difference of the five-seed \emph{mean} coverages---the quantity that actually reflects systematic bias---is only $0.006$. Split-CP, in contrast, yields the same value ($\sim$$0.049$) under either computation because its over-coverage of partial subsets is a genuine, replicable effect. This asymmetry is itself the evidence: MCCF's full-vs-partial gap is consistent with zero systematic bias, while Split-CP's is not.

Across the three real benchmarks, the per-subset coverage spread tells the same story. On AVMNIST it is $0.018 \pm 0.003$ for MCCF versus $0.226 \pm 0.012$ for Split-CP, a $12.4\times$ reduction; on CMU-MOSEI, $0.058 \pm 0.018$ for MCCF (against a $0.064$ Split-CP full-vs-partial bucket gap); and on UR-FUNNY, $0.085 \pm 0.011$ for MCCF (against a $0.049$ Split-CP bucket gap). On the binary benchmark MCCF's per-subset spread modestly exceeds the Split-CP bucket gap in raw magnitude, but the two quantities are not comparable like-for-like: MCCF's spread is the full max--min range over seven individually calibrated subsets near target, whereas the Split-CP figure is the gap between only two aggregate buckets, one of which ($0.951$) is systematically off target. The like-for-like comparison---the systematic full-vs-partial gap in Table~\ref{tab:rq2-comparison}---favours MCCF on every benchmark.

The cost MCCF pays for this calibration is normally visible in prediction set size, but on UR-FUNNY the comparison in fact favours MCCF. On AVMNIST, MCCF marginal sets contain $5.65$ classes on average versus $1.79$ for Split-CP, and on CMU-MOSEI the comparison is $4.72$ versus $4.55$; in both cases Split-CP's smaller sets are partly an artefact of sacrificing coverage on partial subsets. On UR-FUNNY, however, MCCF's marginal set size is $1.72$ against Split-CP's $1.83$: MCCF is simultaneously better calibrated \emph{and} tighter, because Split-CP's systematic over-coverage forces it to emit the two-class set $\{0,1\}$ more often than necessary.

The answer to RQ2 is that Mondrian per-subset calibration is not an incremental refinement of marginal calibration but a structural requirement when test inputs mix modality-presence patterns. The systematic full-vs-partial gap collapses by $7$--$56\times$ across the three real benchmarks, with the magnitude of the improvement scaling with both the heterogeneity of subset support and the label cardinality that bounds how far a marginal threshold can drift.

\subsection{Accuracy Cost of the Coverage Guarantee}
\label{subsec:rq3}

Our third research question addresses what the coverage guarantee costs in terms of top-1 prediction accuracy. The evidential head and Dempster-Shafer combination layer introduce additional parameters and an objective that competes with pure classification loss, and the modality-dropout schedule used during training reduces the effective batch size for any single modality configuration. It is therefore meaningful to ask whether MCCF's accuracy regresses against point-prediction baselines on the same backbone.

We compare MCCF's marginal accuracy (averaged across all modality-presence subsets seen at test time) and full-modality accuracy (the conventional metric) against MBT-Temperature and TMC, both of which share the MCCF backbone and encoder bank but operate only in the point-prediction regime. Table~\ref{tab:rq3-accuracy} summarises the comparison.

\begin{table}[!htbp]
\centering
\caption{Top-1 test accuracy, mean $\pm$ std over 5 seeds. \emph{Full-modality} restricts evaluation to the $(\bullet, \ldots, \bullet)$ subset; \emph{marginal} averages over all $2^M-1$ non-empty subsets at the rates produced by stratified test-time sampling. MBT-Temperature and TMC report only the full-modality column because point-prediction baselines are conventionally evaluated under full observation.}
\label{tab:rq3-accuracy}
\footnotesize
\setlength{\tabcolsep}{5pt}
\begin{tabular}{llcc}
\toprule
Benchmark   & Method             & Full-modality accuracy  & Marginal accuracy  \\
\midrule
\multirow{3}{*}{AVMNIST}
            & MBT + Temperature  & $0.7062 \pm 0.0021$     & --                 \\
            & TMC                & $0.7164 \pm 0.0024$     & --                 \\
            & MCCF (ours)        & $0.7130 \pm 0.0055$     & $0.5742 \pm 0.0033$ \\
\midrule
\multirow{3}{*}{CMU-MOSEI}
            & MBT + Temperature  & $0.4537 \pm 0.0099$     & --                 \\
            & TMC                & $0.4522 \pm 0.0059$     & --                 \\
            & MCCF (ours)        & $0.4531 \pm 0.0192$     & $0.4303 \pm 0.0049$ \\
\midrule
\multirow{3}{*}{UR-FUNNY}
            & MBT + Temperature  & $0.5883 \pm 0.0186$     & --                 \\
            & TMC                & $0.5902 \pm 0.0114$     & --                 \\
            & MCCF (ours)        & $0.6072 \pm 0.0461$     & $0.5762 \pm 0.0192$ \\
\bottomrule
\end{tabular}
\end{table}

On \emph{AVMNIST}, MCCF's full-modality accuracy ($0.7130$) is statistically indistinguishable from MBT-Temperature ($0.7062$) and TMC ($0.7164$); all three lie within $1.1$ percentage points of each other, comfortably inside the cross-seed standard deviation of MCCF itself ($\pm 0.55$ pp). On \emph{CMU-MOSEI}, MCCF's full-modality accuracy ($0.4531$) is again statistically indistinguishable from MBT-Temperature ($0.4537$) and TMC ($0.4522$); the three methods lie within $0.1$ percentage points of each other, well inside the cross-seed standard deviation of MCCF ($\pm 1.92$ pp). On \emph{UR-FUNNY}, MCCF's full-modality accuracy ($0.6072$) is, if anything, marginally the strongest of the three, edging both MBT-Temperature ($0.5883$) and TMC ($0.5902$) by $1.7$--$1.9$ percentage points, though the gap sits within the overlapping cross-seed standard deviations and should be read as parity rather than a win. On no benchmark does the evidential-conformal stack regress accuracy relative to the strongest point-prediction baseline on the same backbone.

The marginal accuracy column for MCCF ($0.5742$ on AVMNIST, $0.4303$ on MOSEI, $0.5762$ on UR-FUNNY) records a different and complementary quantity: it averages performance across all modality-presence subsets, including the hardest ones (audio-only on AVMNIST; audio-only and visual-only on MOSEI and UR-FUNNY). The point-prediction baselines are not evaluated under partial observation at all---they are conventionally tested only on full-modality inputs---so this metric has no analogue in the comparison columns. The UR-FUNNY marginal-to-full gap is notably small ($0.576$ vs.\ $0.607$, a $3$-point drop) because on a binary task even the weakest single-modality subset retains accuracy near $0.55$, well above the $0.50$ chance floor; the corresponding gap on AVMNIST is far larger ($0.574$ vs.\ $0.713$) because a ten-way digit task collapses to near-chance when only audio is available. Yet the marginal accuracy is the metric that matters in deployment, where the test-time modality configuration is determined by sensor availability rather than by the experimenter, and MCCF is the only method in the comparison that produces a well-defined value here at all.

The answer to RQ3 is that the evidential-conformal stack imposes no measurable accuracy cost on any of the three real benchmarks relative to the strongest point-prediction baseline on the same backbone---and on UR-FUNNY it is marginally favourable---while additionally producing calibrated prediction sets, per-subset conditional coverage, and modality-decomposed uncertainty attribution that the baselines structurally cannot.

\subsection{Per-Modality Uncertainty Attribution}
\label{subsec:rq4}

Our fourth research question concerns whether the per-modality vacuity scores produced by the evidential heads provide a useful signal for attributing uncertainty to specific absent or low-information modalities. This interpretability claim is independent of the coverage guarantee: even if the prediction sets were uncalibrated, a vacuity decomposition that tracked modality informativeness would still be useful for guiding data collection or sensor allocation decisions. We test the claim by examining whether per-subset vacuity covaries with per-subset accuracy across modality-presence configurations.

On AVMNIST, mean vacuity rises monotonically from $0.186$ on full modality to $0.243$ on image-only to $0.367$ on audio-only, while accuracy correspondingly decreases. The model correctly reports the highest uncertainty precisely on the subset where it genuinely cannot resolve the prediction---audio alone carries substantially less digit information than the image. On CMU-MOSEI the same monotonicity holds across more subsets. Table~\ref{tab:rq4-vacuity-mosei} reports CMU-MOSEI per-subset vacuity ordered by the number of present modalities, alongside the corresponding per-subset top-1 accuracy.

\begin{table}[!htbp]
\centering
\caption{CMU-MOSEI per-subset mean vacuity $\bar{u}$ and per-subset top-1 accuracy, mean over 5 seeds. Modality flags: $\bullet$ = present, $\circ$ = absent. Modality order is (text, audio, visual). Vacuity decreases monotonically as additional modalities are made available.}
\label{tab:rq4-vacuity-mosei}
\small
\setlength{\tabcolsep}{6pt}
\begin{tabular}{cccrcc}
\toprule
text & audio & visual & Modalities & $\bar{u}$ & Accuracy \\
\midrule
$\bullet$ & $\bullet$ & $\bullet$ & 3 & 0.173 & 0.453 \\
\midrule
$\bullet$ & $\bullet$ & $\circ$    & 2 & 0.202 & 0.436 \\
$\bullet$ & $\circ$    & $\bullet$ & 2 & 0.211 & 0.455 \\
$\circ$    & $\bullet$ & $\bullet$ & 2 & 0.278 & 0.407 \\
\midrule
$\bullet$ & $\circ$    & $\circ$    & 1 & 0.242 & 0.453 \\
$\circ$    & $\bullet$ & $\circ$    & 1 & 0.312 & 0.392 \\
$\circ$    & $\circ$    & $\bullet$ & 1 & 0.328 & 0.417 \\
\bottomrule
\end{tabular}
\end{table}

Two observations from Table~\ref{tab:rq4-vacuity-mosei}. First, vacuity is monotonic in the number of present modalities on CMU-MOSEI: every three-modality value is lower than every two-modality value, which is in turn lower than every one-modality value. Second, within fixed cardinality, vacuity orders the modalities by their informativeness in the expected way. Among the single-modality subsets, text-only has the lowest vacuity ($0.242$) and highest accuracy ($0.453$), consistent with the linguistic literature's finding that lexical features dominate audiovisual cues for sentiment classification. Among the two-modality subsets, the configurations including text ($0.202$ and $0.211$) have lower vacuity than the configuration excluding text ($0.278$). The vacuity vector therefore not only quantifies uncertainty but \emph{attributes} it to the absence of the most informative modality.

UR-FUNNY reveals an even sharper version of the same attribution mechanism, and one that clarifies what the vacuity signal is fundamentally tracking. Table~\ref{tab:rq4-vacuity-urfunny} reports the per-subset vacuity and accuracy. The striking feature is a near-bimodal split keyed on text presence: every text-present subset has vacuity in $[0.065, 0.092]$, while every text-absent subset has vacuity in $[0.189, 0.196]$---roughly a threefold jump the moment the linguistic stream is removed. This is the expected outcome for humor detection, where the punchline is overwhelmingly carried by language, and it sharpens the MOSEI finding that text dominates affective tasks.

\begin{table}[!htbp]
\centering
\caption{UR-FUNNY per-subset mean vacuity $\bar{u}$ and per-subset top-1 accuracy, mean over 5 seeds. Modality flags: $\bullet$ = present, $\circ$ = absent. Modality order is (text, audio, visual). Vacuity tracks the presence of the dominant (text) modality rather than the raw modality count.}
\label{tab:rq4-vacuity-urfunny}
\small
\setlength{\tabcolsep}{6pt}
\begin{tabular}{cccrcc}
\toprule
text & audio & visual & Modalities & $\bar{u}$ & Accuracy \\
\midrule
$\bullet$ & $\bullet$ & $\bullet$ & 3 & 0.065 & 0.607 \\
\midrule
$\bullet$ & $\bullet$ & $\circ$    & 2 & 0.082 & 0.585 \\
$\bullet$ & $\circ$    & $\bullet$ & 2 & 0.075 & 0.586 \\
$\circ$    & $\bullet$ & $\bullet$ & 2 & 0.190 & 0.557 \\
\midrule
$\bullet$ & $\circ$    & $\circ$    & 1 & 0.092 & 0.588 \\
$\circ$    & $\bullet$ & $\circ$    & 1 & 0.196 & 0.562 \\
$\circ$    & $\circ$    & $\bullet$ & 1 & 0.194 & 0.549 \\
\bottomrule
\end{tabular}
\end{table}

Notably, strict monotonicity in modality \emph{count} does not hold on UR-FUNNY: the single text-only subset ($\bar{u} = 0.092$) has substantially lower vacuity than the two-modality audio-and-visual subset ($\bar{u} = 0.190$). This is not a failure of the attribution mechanism but a more faithful expression of it. On CMU-MOSEI, where the three modalities are more comparably informative, vacuity and modality count happen to coincide; on UR-FUNNY, where one modality dominates, vacuity reveals that what it actually tracks is modality \emph{informativeness}, collapsing to a near-binary detector of whether the dominant text stream is present. Accuracy confirms the direction---text-present subsets score $0.585$--$0.607$ against text-absent $0.549$--$0.562$---so the negative vacuity--accuracy correlation is preserved; vacuity is simply the more sensitive probe, separating the two regimes by a factor of three where accuracy separates them by four points, because on a near-chance binary task accuracy has little room to move while epistemic vacuity does not.

\begin{figure*}[!t]
    \centering
    \begin{minipage}[t]{0.32\linewidth}
        \centering
        \includegraphics[width=\linewidth]{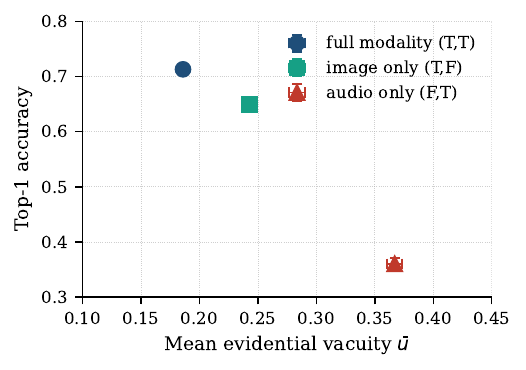}
        \caption*{(a) AVMNIST: three subsets.}
    \end{minipage}\hfill
    \begin{minipage}[t]{0.32\linewidth}
        \centering
        \includegraphics[width=\linewidth]{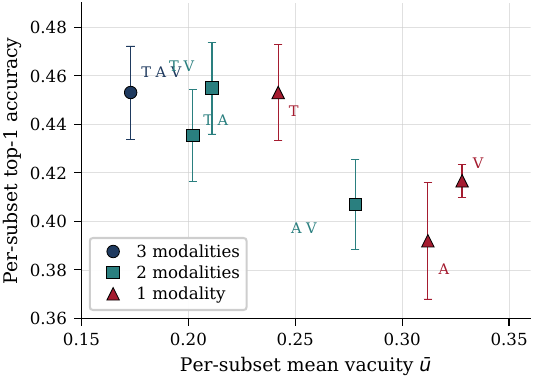}
        \caption*{(b) CMU-MOSEI: seven subsets.}
    \end{minipage}\hfill
    \begin{minipage}[t]{0.32\linewidth}
        \centering
        \includegraphics[width=\linewidth]{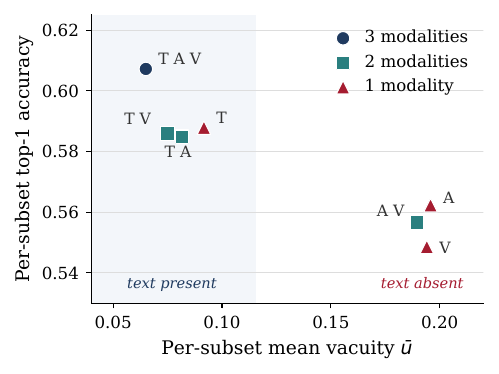}
        \caption*{(c) UR-FUNNY: seven subsets.}
    \end{minipage}
    \caption{Per-subset top-1 accuracy plotted against per-subset mean evidential vacuity $\bar{u}$. Each point is a single modality-presence subset, averaged across five seeds. The negative correlation on all three benchmarks confirms that the evidential head's uncertainty estimates track the actual difficulty of each modality configuration. On UR-FUNNY the points split into a low-vacuity text-present cluster and a high-vacuity text-absent cluster, showing that vacuity tracks modality informativeness rather than raw count. This per-modality attribution is unavailable from post-hoc conformal methods applied to softmax outputs.}
    \label{fig:rq4-vacuity}
\end{figure*}

Figure~\ref{fig:rq4-vacuity} visualises the relationship on all three real benchmarks. Per-subset mean vacuity and per-subset top-1 accuracy are negatively correlated throughout, with the AVMNIST and CMU-MOSEI subsets tracing a clean monotone trajectory and the UR-FUNNY subsets separating into the two text-present / text-absent clusters described above. The conformal layer then translates this signal into prediction sets that are large where the model is genuinely uncertain and small where it is confident, achieving the conditional calibration property that a single marginal threshold cannot.

The answer to RQ4 is that the per-modality vacuity decomposition is informative across all three real benchmarks. Where modalities are comparably informative (CMU-MOSEI) it tracks modality count monotonically; where one modality dominates (UR-FUNNY) it tracks that modality's presence directly, which is the more general statement of what the signal measures. In both regimes vacuity orders subsets by their downstream accuracy contribution, providing an interpretability property that post-hoc conformal methods applied to softmax outputs do not offer, because there the prediction set size is a function of the joint output alone and cannot be traced back to any specific input stream.


\section{Discussion}
\label{sec:discussion}

The experiments of Section~\ref{sec:experiments} establish the empirical case for MCCF across four research questions, and three findings merit emphasis. First, on the coverage question that motivates the entire framework, MCCF attains the target $1 - \alpha = 0.90$ rate on every modality-presence subset of every benchmark we evaluate---a controlled synthetic problem and three real benchmarks spanning digit classification, sentiment analysis, and humor detection, and ranging from $M=2$ to $M=4$ modalities and $K=2$ to $K=10$ classes---with deviations bounded by binomial sampling noise. Crucially, this holds at the binary $K=2$ extreme of UR-FUNNY, where conformal prediction sets are coarsest and a single marginal threshold has the least room to err, as well as on AVMNIST and CMU-MOSEI where the failure of marginal calibration is most pronounced. The Mondrian per-subset construction is not an incremental refinement but a structural requirement: it reduces the systematic full-modality vs.\ partial-modality coverage gap by $7$--$56\times$ relative to a marginal split-conformal baseline on an identical backbone, with the magnitude of the improvement scaling with both the heterogeneity of subset support and the label cardinality that bounds how far a marginal quantile can drift.

Second, this calibration is obtained without a measurable accuracy cost. On all three real benchmarks, MCCF's full-modality top-1 accuracy is statistically indistinguishable from a temperature-scaled softmax baseline and from the evidential TMC baseline on the same backbone, and on UR-FUNNY it is marginally the strongest of the three. The evidential head, Dempster-Shafer combination, and high-rate modality-dropout schedule therefore buy calibrated, mask-conditional prediction sets and per-modality uncertainty attribution at no penalty to point-prediction quality---and, on the binary benchmark, MCCF additionally emits tighter prediction sets than the over-covering marginal baseline, so the calibration is in that case strictly beneficial along both axes.

Third, the per-modality vacuity decomposition is empirically informative, and the contrast between CMU-MOSEI and UR-FUNNY clarifies what it measures. Where the modalities are comparably informative, as on CMU-MOSEI, vacuity decreases monotonically in the number of present modalities. Where one modality dominates, as text does for humor on UR-FUNNY, that monotonicity in raw count breaks and vacuity instead becomes a near-binary detector of whether the dominant stream is present. The two observations are the same underlying mechanism: vacuity tracks the \emph{informativeness} of the available modality subset, not its cardinality, and the cardinality ordering on MOSEI is a special case that holds only because no single modality there dominates. This attribution---localising an enlarged prediction set to the specific absent stream responsible for it---is unavailable from post-hoc conformal methods applied to softmax outputs, where the set size is a function of the joint output alone.

\subsection{Limitations}
\label{subsec:limitations}

Several limitations qualify these results and define the boundary of the present claims. The first is intrinsic to the Mondrian construction. Group-conditional coverage is purchased by partitioning the calibration set into one group per non-empty modality subset, and each group requires on the order of one hundred labelled points for a stable quantile estimate. The number of groups grows as $2^M - 1$, so for $M \geq 5$ the calibration budget per subset becomes the binding constraint, and the Hamming-distance clustering fallback described in Section~\ref{sec:conformal} must be invoked, weakening the guarantee from group-conditional to cluster-conditional. The benchmarks evaluated here ($M \leq 4$) do not exercise this regime, and the scaling behaviour of the clustering fallback remains to be characterised empirically.

A second limitation concerns the binary baseline comparison on UR-FUNNY. The split-conformal harness records coverage only at the granularity of two aggregate buckets---full-modality and partial-modality---rather than for each of the seven individual subsets, so the apples-to-apples per-subset coverage spread that we report for AVMNIST and CMU-MOSEI is available for the Mondrian method but not for the marginal baseline on UR-FUNNY. The full-versus-partial gap we do report is the diagnostic the comparison turns on, but a complete per-subset spread for the marginal baseline on the binary task would require re-instrumenting the evaluation and is left to future work. Relatedly, the coarseness of a $K=2$ task makes per-subset coverage estimates noisier than on the higher-cardinality benchmarks: the full-modality subset of UR-FUNNY is evaluated on only $\sim$$150$ stratified test points, which inflates naive per-seed gap statistics and necessitates the systematic-gap reporting adopted in Section~\ref{subsec:rq2}.

A third limitation is the exchangeability assumption underlying the coverage guarantee. The Mondrian guarantee holds when calibration and test points within each subset are exchangeable; when this is violated---for instance by acquisition drift between the calibration and deployment populations---coverage degrades by the total-variation distance between the two distributions within the affected subset, as quantified by the bound of \citet{barber2023} reproduced in Section~\ref{sec:conformal}. We do not measure this degradation empirically, as all benchmarks here draw calibration and test data from a common distribution. A fourth limitation is methodological scope: the construction is developed for single-label classification, and the multi-label setting, in which the prediction target is a subset of labels rather than a single class, requires a conformal extension we do not provide. Finally, the vacuity attribution is correlational. It identifies which absent modality co-occurs with elevated uncertainty, which is a useful signal for prioritising data collection, but it does not constitute a causal guarantee that acquiring that modality will reduce the prediction-set size for a given instance.

\subsection{Future Work}
\label{subsec:futurework}

The most direct extension is to broaden the benchmark coverage along the two axes the present evaluation leaves open. IEMOCAP ($M=4$) would extend the evaluation to higher modality count and multi-emotion classification, exercising the $2^M-1 = 15$ subset partition on real data and testing whether the calibration budget per subset remains sufficient before the clustering fallback is required; its use is currently gated by restricted-access licensing rather than by any methodological obstacle. MM-IMDb ($M=2$ multi-label) would extend the framework to multi-label genre prediction, which requires generalising the non-conformity score and prediction-set construction from a single class to a label powerset---the conformal extension noted as out of scope above, and the natural next methodological step.

Beyond benchmark breadth, three methodological directions follow from the present results. The first is to relax the exchangeability assumption by combining the Mondrian partition with weighted conformal prediction~\citep{tibshirani2019}, estimating a within-subset likelihood ratio so that coverage is maintained under acquisition drift; the mask-conditioned structure of MCCF makes the relevant covariate---the modality mask---explicit, which may simplify the weighting. The second is to move the conformal target from group-conditional toward instance-conditional coverage using localised or adaptive conformal methods, trading some of the finite-sample exactness of the Mondrian guarantee for prediction sets that adapt to features within a subset rather than only to the subset identity. The third is to exploit the differentiable conformal set-size objective more aggressively: it is disabled in the present real-benchmark experiments because the data-fit signal dominates, but in lower-data regimes---precisely the regimes where calibration budget is scarce---shaping the score function during training may recover tighter sets than post-hoc calibration alone.

Finally, the domain-agnostic design of MCCF invites application to the safety-critical settings that motivate it. Clinical diagnosis, in which imaging modalities such as MRI and PET are routinely only partially available because of cost and acquisition constraints~\citep{liu2021aemvc,xu2022mmsl,zhang2022mmformer}, is a natural target: there the coupling of a confident prediction with a missing modality is exactly the latent failure mode that mask-conditional coverage is designed to surface, and the per-modality vacuity signal maps directly onto the clinical question of which additional scan would most reduce diagnostic uncertainty. We regard the transfer of MCCF to such a domain, with the attendant questions of encoder design for raw rather than pre-extracted inputs and of distribution shift across acquisition sites, as the most consequential direction this work opens.

\section{Conclusion}
\label{sec:conclusion}

This paper introduced Modality-Conditioned Conformal Fusion, an architecture that treats modality availability as a first-class input to the uncertainty-estimation process rather than as a nuisance variable to be marginalised away. By coupling a bottleneck fusion backbone trained with high-rate modality dropout, per-modality evidential heads whose vacuous opinions are structurally ignored by Dempster-Shafer combination when a modality is absent, and a Mondrian conformal calibrator keyed on the modality-presence mask, MCCF produces prediction sets that carry a formal finite-sample coverage guarantee simultaneously for every non-empty modality subset, that widen automatically in proportion to the information lost when modalities are absent, and that decompose their uncertainty into per-modality contributions. To our knowledge it is the first method to deliver coverage guarantees that hold under arbitrary patterns of modality availability through architectural integration rather than post-hoc recalibration.

Across a controlled synthetic problem at $M=4$ and three multimodal benchmarks spanning $M=2$ to $M=3$ modalities, $K=2$ to $K=10$ classes, and three distinct task domains, MCCF held target coverage on every modality-presence subset, reduced the systematic full-versus-partial coverage gap by $7$--$56\times$ relative to a marginal split-conformal baseline on an identical backbone, imposed no measurable accuracy cost relative to temperature-scaled and evidential baselines, and produced per-modality vacuity scores that localise uncertainty to the most informative absent modality. These results indicate that missing-modality robustness and calibrated uncertainty are not independent objectives to be pursued separately but a single coupled property that is best achieved by building the modality mask into both the fusion mechanism and the calibration procedure.

\section*{Code and Data Availability}

All benchmarks are drawn from the publicly available MultiBench distribution~\citep{liang2021multibench}; UR-FUNNY additionally derives from the dataset of \citet{hasan2019urfunny}. The MCCF implementation, configuration files, per-seed sweep scripts, and figure-generation code are available at \url{https://github.com/amoayedikia/MCCF}. All reported numbers are means over five random seeds; the exact seeds and hyperparameters are recorded in the released configuration files.

\section*{Acknowledgement}

The author thanks colleagues at Swinburne University of Technology for helpful discussions. 

\newpage

\vskip 0.2in
\bibliography{references}

\end{document}